\documentclass[11pt, a4paper, logo, copyright]{preprint-asset/kunumi}

\usepackage[authoryear, sort&compress, round]{natbib}
\usepackage[]{mdframed}
\usepackage{anyfontsize}
\usepackage{listings}
\usepackage{hyperref}
\usepackage{url}
\usepackage{graphicx}
\usepackage{mathrsfs} %
\usepackage{etoolbox}
\usepackage{cleveref}
\usepackage{tcolorbox}
\usepackage{tabularx}
\usepackage{colortbl}
\usepackage{booktabs}       %
\usepackage{amsfonts}       %
\usepackage{nicefrac}       %
\usepackage{microtype}      %
\usepackage{subcaption}
\usepackage{multirow}
\usepackage{subcaption}
\usepackage{ragged2e}
\usepackage{enumitem}%
\setlist[itemize]{noitemsep, topsep=0pt}
\usepackage{enumitem,kantlipsum}
\usepackage{tabularx}
\usepackage{ragged2e} %
\usepackage{booktabs} %
\usepackage{xr} 
\usepackage{xcolor}

\newlength\savewidth

\definecolor{baselinecolor}{HTML}{d6eaf8}

\definecolor{mygray}{gray}{0.4}

\definecolor{darkgreen}{rgb}{0, 0.5, 0}

\AtBeginEnvironment{tcolorbox}{\tiny}
\usepackage[utf8]{inputenc} 
\usepackage[T1]{fontenc}    
\usepackage{hyperref}       
\usepackage{url}            
\usepackage{booktabs}       
\usepackage{amsfonts}       
\usepackage{nicefrac}       
\usepackage{microtype}      
\usepackage{xcolor}         
\usepackage{amsmath}
\usepackage{amsthm}

\usepackage{mathtools}
\usepackage{algorithm}
\usepackage{algorithmic}
\usepackage{booktabs}
\usepackage{multirow}
\usepackage{tabularx}
\usepackage{enumitem}
\usepackage{tikz}
\usepackage{amsmath}
\usepackage{xcolor}
\usepackage{nameref}
\usepackage[dvipsnames]{xcolor}
\usetikzlibrary{arrows.meta, positioning, fit, backgrounds, shapes.geometric, calc, decorations.pathreplacing}
\usepackage{siunitx}
\usepackage{graphicx,subcaption}

\newcommand{\method}{\textsc{Slice}\xspace}
\newcommand{\mfgt}{\emph{Fgt}\xspace}
\newcommand{\map}{\emph{AP}\xspace}
\newcommand{\mfp}{\emph{FP}\xspace}
\newcommand{\mgp}{\emph{GP}\xspace}
\newcommand{\mip}{\emph{IP}\xspace}

\usepackage{xspace}

\title{Low-Rank Adapters Initialization via Gradient Surgery for Continual Learning}

\author[*,1]{Joana Pasquali}
\author[*,1]{Ramiro N. Barros}
\author[1]{Arthur S. Bianchessi}
\author[1]{Vinícius Conte Turani}
\author[1]{João~Vitor~Boer~Abitante}
\author[1]{Rafaela Cappelari Ravazio}
\author[1]{Christian Mattjie}
\author[1]{Otávio Parraga}
\author[1]{Lucas~S.~Kupssinskü}
\author[1,2]{Rodrigo~C. Barros}
\affil[1]{MALTA, Machine Learning Theory and Applications Lab, PUCRS, Porto Alegre, Brazil}
\affil[2]{Kunumi Institute, Brazil}
\affil[*]{Equal~contribution}

\newif\ifonecolumn
\onecolumntrue %
\begin{abstract}
\vspace{-1em}
LoRA is widely adopted for continual fine-tuning of Large Language Models due to its parameter efficiency, modularity across tasks, and compatibility with replay strategies. However, LoRA-based continual learning remains vulnerable to catastrophic forgetting, whose severity depends on how successive task gradients interact: when consecutive task gradients conflict, standard adapter initializations channel updates into subspaces that overwrite previously learned directions. We propose \method, a gradient-surgery-based initialization for LoRA adapters in continual learning. \method accumulates gradients from both the current task and a replay buffer of prior tasks, reconciles them through a projection operator, and decomposes the result via truncated SVD to initialize the adapter weights. We evaluate \method on the TRACE benchmark and sequences of Super-NI tasks, including a set of adversarial Super-NI sequences that we construct by mining task pairs with maximally opposing gradients. Compared to vanilla LoRA, LoRA-GA, and LoRAM, \method consistently achieves a better stability-plasticity trade-off, improving Average Performance, Final Performance and Forgetting metrics while preserving General Performance and In Context Performance across both standard and adversarial continual learning sequences.
\end{abstract}

\begin{document}

\maketitle


\section{Introduction}
\label{sec:introduction}

Large language Models (LLMs) are increasingly deployed in settings that demand sequential adaptation to non-stationary task distributions: new domains, evolving instructions, and shifting user requirements arrive over time, and retraining from scratch at each stage is prohibitively expensive. 
Continual learning (CL) provides the natural framework for this regime \cite{shi2025continualsurvey}, but its central pathology---catastrophic forgetting \cite{kirkpatrick2017overcoming}, where optimization on a new task degrades performance on previously learned ones---remains a fundamental obstacle. 

Low-Rank Adaptation (LoRA) \cite{hu2022lora} is the dominant parameter-efficient fine-tuning paradigm, and it offers several advantages for CL: its low parameter count limits the degrees of freedom available for destructive interference, its modularity enables task-specific adapter storage and composition, and its compatibility with replay and regularization strategies makes it suitable for CL \cite{liang2024inflora, liu2024learningattentionalmixtureloras, wang2023olora, xiong2026oplora}. 

In CL, when the gradient of a new task conflicts with the gradients of previously learned tasks---that is, when their Frobenius inner product is negative---any descent step that improves current-task performance degrades prior-task performance. 
The adapter initialization determines the subspace within which all subsequent optimization occurs, and standard initialization schemes are blind to this conflict structure \cite{meng2024pissa}. 
Vanilla LoRA initializes $A$ with random Gaussian entries and $B = 0$, placing the adapter in a direction uncorrelated with any task objective. 
Spectral methods such as PiSSA~\cite{meng2024pissa} and MiLoRA~\cite{wang2024milora} derive initializations from the singular structure of the pretrained weights, capturing general representational directions but encoding no task-specific signal. 
LoRA-GA \cite{wang2024loraga} takes a step toward task awareness by initializing adapters via SVD of the fine-tuning-task gradient. 
However, none of these methods account for previously learned tasks: LoRA-GA's initialization is optimal for single-task adaptation but may push the adapter into a subspace that overwrites prior knowledge whenever the current and previous task gradients conflict. The initialization stage thus represents a consequential but underexploited intervention point for CL.

We propose \method (Gradient-\textbf{S}urgery-based \textbf{L}ow-rank \textbf{I}nitialization for \textbf{C}ontinual l\textbf{E}arning), a method that initializes LoRA adapters in a subspace that is simultaneously aligned with the current-task objective and minimally destructive to previously acquired task knowledge. 

To better evaluate \method, we introduce NI-SEQ-OPPOSITE, three adversarial 5-task sequences where tasks were mined by exhaustive combinatorial search over $\binom{46}{5}$ candidate subsets from the Super-NI task pool \cite{wang-etal-2022-super} to minimize mean pairwise gradient cosine similarity. 
Existing Super-NI task sequences \cite{jiang2025unlocking} are constructed by grouping tasks according to output type (classification, generation, or mixed) without any explicit criterion linking sequence composition to gradient interference. 

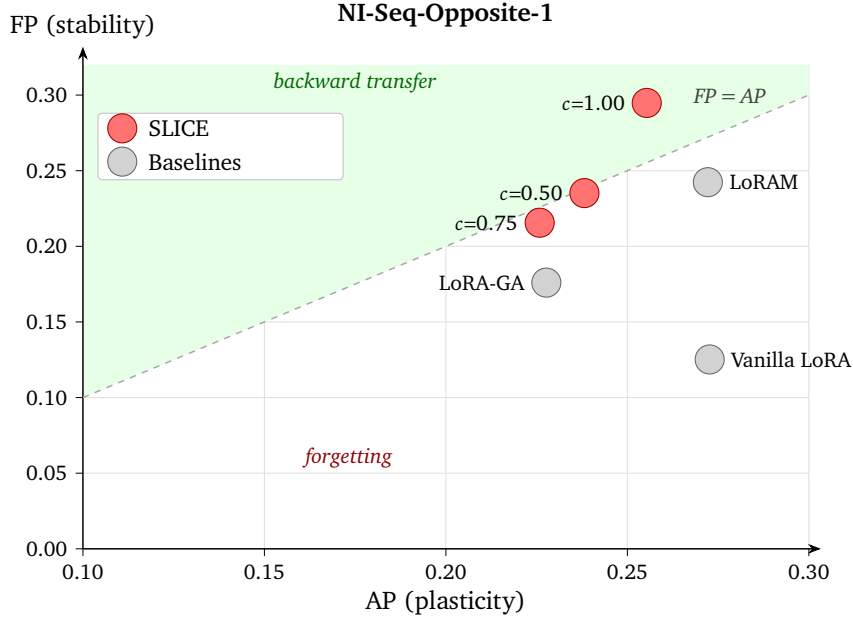
\begin{figure}[t]
\centering



\centering
\scalebox{0.80}{\begin{tikzpicture}[
    >=Stealth,
    font=\normalsize,
    slice/.style ={circle, draw=red!65!black,  line width=0.5pt, fill=red!55,
                   minimum size=4.8mm, inner sep=0pt},
    base/.style  ={circle, draw=black!60,      line width=0.5pt, fill=gray!35,
                   minimum size=4.8mm, inner sep=0pt},
    lbl/.style   ={font=\small, inner sep=1pt},
    lblbold/.style={font=\small\bfseries, inner sep=1pt},
]
\fill[green!10] (0, 2.5) -- (12, 7.5) -- (12, 8) -- (0, 8) -- cycle;
\draw[->, line width=0.7pt] (0,0) -- (12.2, 0);
\node[font=\large, below] at (6, -0.55) {AP (plasticity)};
\draw[->, line width=0.7pt] (0,0) -- (0, 8.3)
    node[above, font=\large] {FP (stability)};
\foreach \x/\v in {0/0.10, 3/0.15, 6/0.20, 9/0.25, 12/0.30}{
    \draw (\x, 0) -- (\x, -0.12);
    \node[below, font=\small] at (\x, -0.12) {\v};
}
\foreach \y/\v in {0/0.00, 1.25/0.05, 2.5/0.10, 3.75/0.15, 5/0.20, 6.25/0.25, 7.5/0.30}{
    \draw (0, \y) -- (-0.12, \y);
    \node[left, font=\small] at (-0.12, \y) {\v};
}
\begin{scope}[on background layer]
    \foreach \x in {3, 6, 9, 12}
        \draw[gray!25, line width=0.3pt] (\x, 0) -- (\x, 8);
    \foreach \y in {1.25, 2.5, 3.75, 5, 6.25, 7.5}
        \draw[gray!25, line width=0.3pt] (0, \y) -- (12, \y);
\end{scope}
\draw[gray!70, dashed, line width=0.7pt] (0, 2.5) -- (12, 7.5)
    node[pos=0.95, above left=-1pt, font=\small\itshape, gray!50!black]
    {FP\,=\,AP};
\node[font=\small\itshape, green!40!black, align=center]
    at (4.5, 7.7) {backward transfer};
\node[font=\small\itshape, red!55!black, align=center]
    at (4.4, 1.5) {forgetting};
\node[slice] (P2) at (9.32,  7.37) {};
\node[slice] (s4)  at (8.29, 5.88) {};
\node[slice] (s6)  at (7.55, 5.39) {};
\node[base]  (b1)  at (10.36, 3.13) {};
\node[base]  (b2)  at (10.33, 6.06) {};
\node[base]  (b3)  at (7.66,  4.40) {};
\node[lblbold, anchor=east, xshift=-2pt] at (P2.west) {$c{=}1.00$};
\node[lbl, anchor=west, xshift=2pt]  at (b2.east) {LoRAM};
\node[lbl, anchor=west, xshift=2pt]  at (b1.east) {Vanilla LoRA};
\node[lbl, anchor=east, xshift=-2pt] at (b3.west) {LoRA-GA};
\node[lbl, anchor=east, xshift=-2pt]  at (s4.west) {$c{=}0.50$};
\node[lbl, anchor=east, xshift=-2pt]  at (s6.west) {$c{=}0.75$};
\begin{scope}[shift={(0.4, 5.6)}]
    \draw[rounded corners=2pt, fill=white, draw=gray!50, line width=0.4pt]
        (-0.15, 0.50) rectangle (3.9, 1.60);
    \node[slice] at (0.25, 1.35) {};
    \node[anchor=west, font=\normalsize] at (0.55, 1.35) {SLICE};
    \node[base] at (0.25, 0.80) {};
    \node[anchor=west, font=\normalsize] at (0.55, 0.80) {Baselines};
\end{scope}
\node[font=\large\bfseries, anchor=south] at (6, 8.5)
    {NI-Seq-Opposite-1};
\end{tikzpicture}}
\par\vspace{1mm}

\caption{
Stability--plasticity trade-off on NI-Seq-Opposite-1. FP (stability) vs.\ AP (plasticity) for \method and baselines under variance-matched rescaling \cite{zhang2025primacy}. The dashed line denotes FP = AP; the shaded region indicates positive backward transfer. 
}
\label{fig:slice-trade-off}
\end{figure}

We evaluate \method on the TRACE benchmark \cite{wang2023tracecomprehensivebenchmarkcontinual} and on both standard (G1, G2) and adversarial (NI-SEQ-OPPOSITE) Super-NI sequences. 
\method is compared to three baseline initializations: vanilla LoRA \cite{hu2022lora}, LoRAM \cite{zhang2025primacy}, and LoRA-GA \cite{wang2024loraga}. 
All comparisons employ variance-matched magnitude rescaling to control for the confound identified by Zhang et al.\ \cite{zhang2025primacy}, whereby apparent initialization gains are attributable to implicit magnitude amplification rather than subspace quality. 
Our results show that \method consistently improves Final Performance and reduces Forgetting while incurring only marginal reductions in General Performance. 
Figure \ref{fig:slice-trade-off} shows that all \method variants ($c \in \{0.50, 0.75, 1.00\}$ consistently achieve higher stability in comparison to all baselines at similar plasticity for NI-Seq-Opp1.\footnote{Code available at: \url{https://github.com/RamiroNB/slice}}


\section{Problem Formulation}
\label{sec:cl}

\paragraph{Continual learning setup.}
Let $\{\mathcal{T}_1, \dots, \mathcal{T}_T\}$ denote a sequence of tasks with data distributions $\mathcal{D}_1, \dots, \mathcal{D}_T$, and let $\theta$ parametrizes a model trained sequentially over them. 
At task $\mathcal{T}_t$, the learner is updated on $\mathcal{D}_t$ alone but is evaluated on all $\mathcal{D}_i$ with $i \le t$. Two opposing population losses characterize this regime---the current-task (plasticity) loss:
\begin{equation}\label{eq:loss-cur}
    \mathcal{L}_{\text{cur}}(\theta)
    \;:=\; \mathbb{E}_{(x,y)\sim\mathcal{D}_t}
        \!\left[\mathcal{L}_t(\theta; x, y)\right],
\end{equation}
and the previous tasks (stability) loss:
\begin{equation}\label{eq:loss-prev}
    \mathcal{L}_{\text{prev}}(\theta)
    \;:=\; \sum_{i=1}^{t-1}
        \mathbb{E}_{(x,y)\sim\mathcal{D}_i}
        \!\left[\mathcal{L}_i(\theta; x, y)\right].
\end{equation}
Sequential fine-tuning only optimizes $\mathcal{L}_{\text{cur}}$ while control over $\mathcal{L}_{\text{prev}}$ is implicit. 
Catastrophic forgetting occurs when the latter grows uncontrollably.


\paragraph{Data access.}
We assume $\bigcup_{i<t}\mathcal{D}_i$ remains accessible for sampling on demand. 
This is strictly weaker than the rehearsal regime~\citep{buzzega2020der, chaudhry2018agem, chaudhry2019er}: \method draws a fresh sample $\mathcal{D}_{\text{prev}} \subseteq \bigcup_{i<t}\mathcal{D}_i$ at initialization, uses it once, and discards it; no persistent memory is replayed during training.

\paragraph{Per-layer gradients.}
For each target weight matrix $W_l$, the two losses induce gradients
\begin{equation}\label{eq:grads-pop}
    G_{\text{cur}} \;:=\; \nabla_{W}\, \mathcal{L}_{\text{cur}}(\theta),
    \qquad
    G_{\text{prev}}\;:=\; \nabla_{W}\, \mathcal{L}_{\text{prev}}(\theta),
\end{equation}
estimated from finite samples of $\mathcal{D}_t$ and $\mathcal{D}_{\text{prev}}$. 
The direction of both gradients provide conflicting information. When $\langle G_{\text{cur}}, G_{\text{prev}} \rangle_F < 0$, a descent step on $\mathcal{L}_{\text{cur}}$ also increases $\mathcal{L}_{\text{prev}}$; when the inner product is negative, the two objectives are incompatible and any improvement on the current task is paid for in stability. 
\method operates directly on this geometry, reconciling $G_{\text{cur}}$ and $G_{\text{prev}}$ into a conflict-free direction prior to adapter construction.
\section{Gradient-Surgery Low-Rank Initialization}
\label{sec:slice}
We introduce \method, a method for initializing low-rank adapters in a subspace aligned with the current-task objective while minimizing interference with previously learned tasks.

In a fine-tuning with LoRA~\citep{hu2022lora}, each target weight matrix $W^{(l)}$ is modified as ${W}' = W_0 + B A$, with $B \in \mathbb{R}^{d_{\text{out}} \times r}$ and $A \in \mathbb{R}^{r \times d_{\text{in}}}$ for rank $\smash[t]{r \ll \min(d_{\text{out}}, d_{\text{in}})}$.

Standard practice initializes $A$ randomly and $B = 0_{d_{\text{out} \times r}}$, making the initial adapter direction agnostic to both the current task $\mathcal{T}_t$ and the previously learned tasks $\{\mathcal{T}_i\}_{i<t}$. 
\method initializes weights in a \emph{conflict-free} update direction by performing a four-stage procedure detailed in Algorithm \ref{alg:slice}.

\begin{figure}[t]
    \resizebox{\textwidth}{!}{\definecolor{slicered}{HTML}{C0392B}
\definecolor{slicewash}{HTML}{FDEDEC}
\definecolor{boxgray}{HTML}{94A3B8}

\begin{tikzpicture}[
    >={Stealth[length=2.5mm, width=2.0mm]},
    font=\Large,
    dbox/.style={
        draw=boxgray, line width=0.5pt, rounded corners=2pt,
        fill=white, minimum width=14mm, minimum height=10mm, inner sep=3pt
    },
    gbox/.style={
        draw=boxgray, line width=0.5pt, rounded corners=2pt,
        fill=white, minimum width=18mm, minimum height=10mm, inner sep=3pt
    },
    psi/.style={
        draw=slicered, line width=0.9pt, rounded corners=4pt,
        fill=slicewash,
        minimum width=36mm, minimum height=22mm,
        align=center, inner sep=5pt
    },
    gtil/.style={
        draw=slicered, line width=0.6pt, rounded corners=2pt,
        fill=slicewash,
        minimum width=16mm, minimum height=10mm, inner sep=3pt
    },
    svd/.style={
        draw=boxgray, line width=0.5pt, rounded corners=2pt,
        fill=white, minimum width=22mm, minimum height=14mm,
        align=center, inner sep=3pt
    },
    phibox/.style={
        draw=boxgray, line width=0.5pt, rounded corners=2pt,
        fill=white, minimum width=34mm, minimum height=10mm, inner sep=3pt, align=center
    },
    magbox/.style={
        draw=boxgray, line width=0.5pt, rounded corners=2pt,
        fill=white,
        minimum width=34mm, minimum height=22mm,
        align=center, inner sep=4pt
    },
    obox/.style={
        draw=boxgray, line width=0.5pt, rounded corners=2pt,
        fill=white, minimum width=28mm, minimum height=10mm, inner sep=3pt
    },
    fl/.style={->,  line width=0.55pt, color=boxgray},
    sl/.style={->,  line width=0.6pt,  color=slicered},
    el/.style={font=\Large, inner sep=1.5pt, color=black!40},
]

\def\ytop{0.75}
\def\ybot{-0.75}

\node[dbox] (DA) at (-0.1,  \ytop) {{\LARGE $\mathcal{D}_{cur}$}};
\node[dbox] (DP) at (0.0,  \ybot) {{\LARGE ${\mathcal{D}}_{prev}$}};
\node[gbox] (GA) at (3.0,  \ytop) {{\LARGE $G_{cur}\vphantom{^{(l)}}$}};
\node[gbox] (GP) at (3.0,  \ybot) {{\LARGE $G_{prev}\vphantom{^{(l)}}$}};

\draw[fl] (DA) -- node[el, above] {{\Large $\nabla\!\mathcal{L}$}} (GA);
\draw[fl] (DP) -- node[el, above] {{\Large $\nabla\!\mathcal{L}$}} (GP);

\node[psi]  (PSI)  at (6.3,  0) {%
    {\LARGE $\psi\!\bigl(G_A,\; G_P\bigr)$}\\[5pt]
    {\Large PCGrad-based}%
};
\node[gtil] (GTIL) at (9.4, 0) {{\LARGE $\tilde{G}_{cur}\vphantom{^{(l)}}$}};

\draw[fl] (GA.east) -- (PSI.west |- GA);
\draw[fl] (GP.east) -- (PSI.west |- GP);
\draw[sl] (PSI)     -- (GTIL);

\node[svd]  (SVD)  at (11.75, 0) {%
    \textbf{SVD}\\[1pt]{\Large $U\Sigma V^{\!\top}$}%
};
\node[phibox] (PB) at (15.25, \ytop) {{\LARGE $\Phi_B^{(l)} = U_{:,\,:r}$ }};
\node[phibox] (PA) at (15.25, \ybot) {{\LARGE $\Phi_A^{(l)} = V_{r:2r,\,:}^{\top}$}};

\draw[fl] (GTIL) -- (SVD);
\draw[fl, rounded corners=2pt] ($(SVD.north east)!1/3!(SVD.south east)$) -- ++(0.35,0) |- (PB.west);
\draw[fl, rounded corners=2pt] ($(SVD.north east)!2/3!(SVD.south east)$) -- ++(0.35,0) |- (PA.west);

\node[magbox] (MAG) at (19.5, 0)   {%
    \textbf{MagScale}\\[3pt]
    {\Large $\beta = (\eta_r\, \eta_{\text{var}})^{1/4}$}\\[3pt]
    {\large match variance to $W_l$}%
};
\node[obox] (BL) at (23.4, \ytop) {{\LARGE $B_l = \beta\,\Phi_B^{(l)}$}};
\node[obox] (AL) at (23.4, \ybot) {{\LARGE $A_l = \beta\,\Phi_A^{(l)}$}};

\draw[fl] (PB.east) -- (MAG.west |- PB);
\draw[fl] (PA.east) -- (MAG.west |- PA);
\draw[fl] (MAG.east |- BL) -- (BL.west);
\draw[fl] (MAG.east |- AL) -- (AL.west);

\end{tikzpicture}}
    \caption{\textbf{\method initialization pipeline.} $\mathcal{D}_{cur}$ and ${\mathcal{D}}_{prev}$ denote the current-task data and prior-task replay buffer; $G_{cur}$, $G_{prev}$ their gradients; $\psi$ is the projection; $\tilde{G}_{cur}$ the reconciled gradient; $\smash[t]{\Phi_B^{(l)}}$, $\smash[t]{\Phi_A^{(l)}}$ the low-rank subspaces; and $\beta = (\eta_r\,\eta_{\mathrm{var}})^{1/4}$ the magnitude-scaling coefficient.}
    \label{fig:pipeline}   
\end{figure}
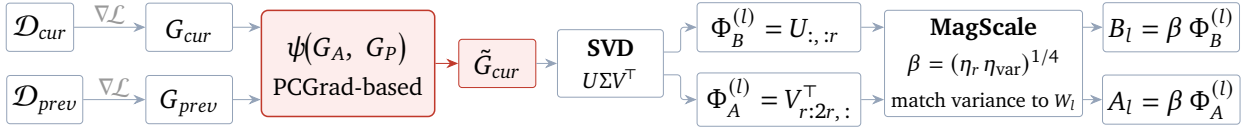

\paragraph{Stage 1: Gradient Estimation.}\label{stg:grad_estimation} We estimate the gradient matrices for the current task and for previously seen task data by accumulating over mini-batches:
\begin{equation}\label{eq:grad-accum}
    G_{\text{cur}} = \frac{1}{S_{\text{cur}}}
        \sum_{s=1}^{S_{\text{cur}}}
        \nabla_{W} \mathcal{L}_{\text{cur}}(\theta;\, \mathcal{B}_s^{\text{cur}}),
    \qquad
    G_{\text{prev}} = \frac{1}{S_{\text{prev}}}
        \sum_{s=1}^{S_{\text{prev}}}
        \nabla_{W} \mathcal{L}_{\text{prev}}(\theta;\, \mathcal{B}_s^{\text{prev}})
\end{equation}
where $\mathcal{B}_s^{\text{cur}} \sim \mathcal{D}_t$ and $\mathcal{B}_s^{\text{prev}} \sim \mathcal{D}_{\text{prev}}$ are mini-batches sampled from the current-task data and from the previous-tasks sample, respectively, and $S_{\text{cur}}$, $S_{\text{prev}}$ are the number of accumulation steps. 
Notably, $S_{\text{cur}}$ and $S_{\text{prev}}$ are small (typically tens of steps), requiring only a lightweight forward--backward pass over a small sample of data from the current task and from $\mathcal{D}_{\text{prev}}$.

\paragraph{Stage 2: Gradient Reconciliation.}\label{stg:grad_recon} We reconcile the current-task gradient with the previous-tasks gradient through a projection operator $\psi$:
\begin{equation}\label{eq:projection}
    \tilde{G}_{\text{cur}}=
    \psi\!\left(G_{\text{cur}},\; G_{\text{prev}}\right)
\end{equation}

The role of $\psi$ is to return an update direction that pursues the current-task (plasticity) objective orthogonal to the previous-tasks (stability) objective.

We employ a generalization of PCGrad parameterized by $c \in [0,1]$.
At $c=1$ the formula recovers standard PCGrad; at $c=0$ no correction is applied, and $\psi$ returns $G_{\text{cur}}$ unchanged. 
The $\min(\cdot,0)$ term is the non-linearity: the correction is active only when the two gradients are conflicting ($\langle G_{\text{cur}}, G_{\text{prev}}\rangle_F < 0$), leaving the gradient unchanged otherwise ($\tilde{G}_{\text{cur}} = G_{\text{cur}}$).

\begin{equation}\label{eq:projection-pc-grad}
    \psi_{\text{PCGrad}}\!\left(G_{\text{cur}}, G_{\text{prev}},c\right) =
    G_{\text{cur}} - c
    \frac{\ \min\bigl(\langle G_{\text{cur}},\; G_{\text{prev}} \rangle_F,\,0\bigr)\ }
         {\| G_{\text{prev}} \|_F^2}\;
    G_{\text{prev}},
\end{equation}



\paragraph{Stage 3: Low-Rank Decomposition.}\label{stg:low_rank_decomp} We decompose the projected gradient via truncated SVD into matrices that will form the adapter initialization. For each target module $l$, let $\smash[t]{\tilde{G}_{\text{cur}}^{(l)} = U \Sigma V^\top}$ be the singular value decomposition. 
We set
\begin{equation}\label{eq:svd-init}
   \Phi_B^{(l)}  = U_{:, :r}^{(l)}, \qquad \Phi_A^{(l)} = (V_{:, r:2r}^{(l)})^\top
\end{equation}
where $U_{:, :r}^{(l)} \in \mathbb{R}^{d_{\text{out}}^{(l)} \times r}$ denotes the first $r$ left singular vectors and $V_{:, r:2r}^{(l)} \in \mathbb{R}^{d_{\text{in}}^{(l)} \times r}$ denotes
the $(r{+}1)$-th through $2r$-th right singular vectors. We acknowledge that this particular factorization is not immediately intuitive---one might naturally expect both $\Phi_B^{(l)}$ and $\Phi_A^{(l)}$ to be initialized from the leading singular components. The choice to draw $\Phi_A^{(l)}$ from the $(r{+}1)$-th through $2r$-th right singular vectors is a deliberate design decision that, while theoretically non-obvious, has been empirically validated in prior work. LoRA-GA~\cite{wang2024loraga} adopts this factorization and reports strong results, demonstrating that it yields effective adapter initializations. Furthermore, LoRAM~\cite{zhang2025primacy} reproduces the LoRA-GA initialization as one of its baselines and similarly obtains strong performance, further corroborating its effectiveness. A complete theoretical explanation for this behavior remains elusive, though it suggests that factors beyond pure approximation quality---such as the optimization landscape induced by the initialization---may play crucial roles in determining optimization behavior. Following these empirical findings, we adopt the same factorization.

\paragraph{Stage 4: Magnitude Rescaling.}\label{stg:mag_rescaling}
Following the magnitude gain initialization~\citep{zhang2025primacy}, we rescale the variance of the effective low-rank reconstruction $B^{(l)}_0A^{(l)}_0$ to match the pretrained weights scale. 
Let
\begin{equation}
\label{eq:beta_pt_1}
\sigma_W^2 \;=\; \mathrm{Var}(W_0^{(l)}),\qquad
\sigma_{BA}^2 \;=\; \mathrm{Var}(\Phi_B^{(l)}\Phi_A^{(l)})
\end{equation}
define the variance ratio $\eta_{\text{var}} = \sigma_W^2 / \sigma_{BA}^2$ and the rank-dependent factor $\eta_r = \log_{m}(r)$ where $m=\min(d_{\text{out}},d_{\text{in}})$. Then

\begin{equation}
\label{eq:beta}
B^{(l)} _0\;=\; \beta \cdot \Phi_B^{(l)}, \quad
A^{(l)}_0 \;=\; \beta \cdot \Phi_A^{(l)}, \quad
\text{where} \quad \beta \;=\; \Big(\eta_r \, \eta_{\text{var}}\Big)^{1/4}.
\end{equation}

Consequently $\mathrm{Var}(B^{(l)}_0A^{(l)}_0)$ is aligned with $\mathrm{Var}(W_0^{(l)})$ up to the prescribed rank correction.

\paragraph{Integration.} The resulting $(A^{(l)}_0, B^{(l)}_0)$ pairs replace the default LoRA initialization. Like other non-zero LoRA initialization methods, such as LoRA-GA and LoRAM, \method requires weight absorption, as described in Appendix~\ref{app:weight-absorption}.
\method is agnostic to the downstream training algorithm: the initialized adapters can be fine-tuned with standard objectives, replay-augmented losses, or regularization-based methods. 
The computational overhead is limited to a gradient accumulation pass over small samples of $\mathcal{D}_t$ and $\mathcal{D}_{\text{prev}}$, followed by one SVD per target module. Figure \ref{fig:pipeline} depicts \method initialization.


\begin{algorithm}[t]
\caption{\method: Gradient-Surgery Low-Rank Initialization}
\label{alg:slice}
\begin{algorithmic}[1]
\REQUIRE Model parameters $\theta = \{W_l\}_{l=1}^{L}$, current-task data
$\mathcal{D}_{cur}$, previous-tasks data sample $\mathcal{D}_{\text{prev}}$,
target modules $\mathcal{T}$, rank $r$, accumulation steps
$S_{\text{cur}}$, $S_{\text{prev}}$, reconciliation operator
$\psi \in \{\psi_{\text{PCGrad}}$\};
\ENSURE Initialized adapter pairs $\{(B^{(l)}_0,A^{(l)}_0)\}_{l \in \mathcal{T}}$

\STATE \textit{\nameref{stg:grad_estimation}}

\STATE $G_{\text{cur}} \leftarrow \frac{1}{S_{\text{cur}}}
        \sum_{s=1}^{S_{\text{cur}}}
        \nabla_{W} \mathcal{L}_{\text{cur}}(\theta;\, \mathcal{B}_s^{\text{cur}})$
        \quad \textit{\ \ \ \ \ // $\mathcal{B}_s^{\text{cur}} \sim \mathcal{D}_{cur}$}
\STATE $G_{\text{prev}} \leftarrow \frac{1}{S_{\text{prev}}}
            \sum_{s=1}^{S_{\text{prev}}}
            \nabla_{W} \mathcal{L}_{\text{prev}}(\theta;\, \mathcal{B}_s^{\text{prev}})$
            \quad \textit{// $\mathcal{B}_s^{\text{prev}} \sim \mathcal{D}_{\text{prev}}$}

\STATE

\STATE \textit{\nameref{stg:grad_recon}} (Eq.~\ref{eq:projection-pc-grad})
\STATE $\tilde{G}_{\text{cur}} \leftarrow
            \psi\!\left(G_{\text{cur}},\; G_{\text{prev}},
            c\right)$

\STATE

\FOR{each target module $l \in \mathcal{T}$}
    \STATE \textit{\nameref{stg:low_rank_decomp}}
    \STATE $U, \Sigma, V \leftarrow \mathrm{SVD}(\tilde{G}_{\text{cur}}^{(l)})$
    \STATE $\Phi_B^{(l)} \leftarrow U_{:, :r}$
    \STATE $\Phi_A^{(l)} \leftarrow (V_{:, r:2r})^\top$

\STATE

    \STATE \textit{\nameref{stg:mag_rescaling}} (Eq.~\ref{eq:beta})
    \STATE $\beta \leftarrow \texttt{MagnitudeScale}\left(\Phi_A^{(l)},\Phi_B^{(l)},W^{(l)}\right)$
    \STATE $A^{(l)}_0 \leftarrow \beta\,\Phi_A^{(l)}$
    \STATE $B^{(l)}_0\leftarrow \beta\,\Phi_B^{(l)}$
\ENDFOR
\STATE

\RETURN $(B_0, \ A_0)$
\end{algorithmic}
\end{algorithm}

\section{Empirical Analysis}
\label{sec:experiments}

We evaluate \method against vanilla initialization and two strong initialization baselines---LoRAM and LoRA-GA---and compare the reconciliation operator $\psi_{\text{PCGrad}}$ (Eq.~\ref{eq:projection-pc-grad}) across $c \in \{0.5, 0.75, 1.0\}$. 
Further details on the experimental setup are described in Appendix \ref{app:experimental}.

\paragraph{Datasets.}We evaluate \method on two settings. 
First, following previous work \citep{jiang2025unlocking,wang2023olora}, we construct task sequences from Super-NaturalInstructions (Super-NI)~\citep{wang-etal-2022-super}, restricting our selection to {purely generative} tasks (i.e., free-form text generation, excluding classification). 
Second, we evaluate on \textsc{Trace}~\citep{wang2023tracecomprehensivebenchmarkcontinual}, a benchmark explicitly designed to stress-test CL methods, comprising diverse task families with documented interference patterns between sequential fine-tuning stages. 
Both settings satisfy the core desiderata for CL evaluation: they expose models to non-stationary task distributions and jointly measure forward transfer and catastrophic forgetting.
The Super-NI sequences used in prior work~\citep{jiang2025unlocking, wang2023olora} are constructed by grouping tasks according to surface-level criteria---output type or NLP task category---without any explicit criterion linking sequence composition to gradient interference. However, sequences whose consecutive tasks exhibit opposing gradients pose a strictly harder continual learning problem, as every descent step on the current task degrades prior-task performance to first order. To provide a controlled stress test, we introduce the \textsc{NI-Seq-Opposite}: 5-task sequences mined by exhaustive combinatorial search over $\binom{46}{5}$ candidate subsets to minimize mean pairwise gradient cosine similarity. Combined with evaluation on TRACE and the standard G1/G2 sequences, the resulting protocol spans the full spectrum of inter-task interference regimes, exposing differences between methods that gradient-aligned sequences leave latent. We detail the construction procedure and per-task gradient alignment in Appendix~\ref{app:opp-sequences}.

\paragraph{Evaluation.} Following the standard CL evaluation framework, we construct a results matrix $\mathbf{R} \in \mathbb{R}^{T \times T}$ where entry $R_{i,j}$ denotes the performance of the model on task $j$ after completing training on task $i$, evaluated on each task's held-out split. 
From this matrix, we derive four CL metrics. 
Average Performance (\map) is the mean of the diagonal entries $R_{i,i}$, capturing task-specific performance measured immediately after each task is trained, thereby reflecting the model's peak retention before any subsequent forgetting. 
Final Performance (\mfp) is the mean of the last row, $R_{T,:}$, measuring performance across all tasks after the full sequence has been trained and thus capturing how much task knowledge is retained at the end of training. 
Forgetting (\mfgt) is the difference \map$-$\mfp, quantifying the average performance degradation induced by subsequent task training. 
All three metrics—\map, \mfp, and \mfgt—use the evaluation splits of the CL task sequence itself (e.g., Super-NaturalInstructions or TRACE), with generation-based evaluation scoring each task by exact match or ROUGE-L depending on the task type. 
In addition, we evaluate general language model capabilities on a fixed set of held-out benchmarks never encountered during training.
General Performance (\mgp) is the mean zero-shot accuracy across HellaSwag, CommonsenseQA, and Alpaca (scored via ROUGE-L), evaluated after the final training stage, measuring the preservation of broad language understanding.
In-context Performance (\mip) evaluates the same three benchmarks, plus BBH Object Counting with few-shot prompting (5-shot for most tasks, 3-shot for BBH), and measures the model's ability to leverage in-context demonstrations after continual training. Detailed computation of the metrics \map, \mfp, \mgp, \mip, and \mfgt is presented in Appendix~\ref{appendix:cl-metrics}.

\begin{table*}[!t]
\centering
\caption{CL metrics across all sequences: baseline values with $\Delta$ for each \method variant (rank 64)}
\label{tab:iclr_delta_all_r64}
\setlength{\tabcolsep}{2pt}
\resizebox{\linewidth}{!}{%
\begin{tabular}{l|ccc|ccc|ccc|ccc|ccc|ccc}
\toprule
 & \multicolumn{3}{c}{G1} & \multicolumn{3}{c}{G2} & \multicolumn{3}{c}{TRACE} & \multicolumn{3}{c}{Opp1} & \multicolumn{3}{c}{Opp2} & \multicolumn{3}{c}{Opp3} \\ \cmidrule(l){2-19}
Method & \map $\uparrow$ & \mfp $\uparrow$ & \mfgt $\downarrow$ & \map $\uparrow$ & \mfp $\uparrow$ & \mfgt $\downarrow$ & \map $\uparrow$ & \mfp $\uparrow$ & \mfgt $\downarrow$ & \map $\uparrow$ & \mfp $\uparrow$ & \mfgt $\downarrow$ & \map $\uparrow$ & \mfp $\uparrow$ & \mfgt $\downarrow$ & \map $\uparrow$ & \mfp $\uparrow$ & \mfgt $\downarrow$ \\
\midrule
Vanilla LoRA & 10.29 & 11.06 & -0.77 & 18.83 & 9.04 & 9.79 & 10.21 & 10.28 & -0.06 & 27.27 & 12.53 & 14.74 & 26.50 & 25.10 & 1.40 & 26.32 & 4.32 & 22.00 \\
\textit{\method ($c{=}0.50$)} & \textcolor{BrickRed}{-0.75} & \textcolor{BrickRed}{-2.03} & \textcolor{BrickRed}{+1.28} & \textcolor{ForestGreen}{\textbf{+16.87}} & \textcolor{ForestGreen}{\textbf{+22.71}} & \textcolor{ForestGreen}{\textbf{-5.84}} & \textcolor{ForestGreen}{\textbf{+3.04}} & \textcolor{ForestGreen}{\textbf{+3.35}} & \textcolor{ForestGreen}{\textbf{-0.31}} & \textcolor{BrickRed}{-3.44} & \textcolor{ForestGreen}{\textbf{+10.98}} & \textcolor{ForestGreen}{\textbf{-14.43}} & \textcolor{BrickRed}{-0.98} & \textcolor{BrickRed}{-10.72} & \textcolor{BrickRed}{+9.75} & \textcolor{BrickRed}{-3.50} & \textcolor{ForestGreen}{\textbf{+5.69}} & \textcolor{ForestGreen}{\textbf{-9.18}} \\
\textit{\method ($c{=}0.75$)} & \textcolor{BrickRed}{-0.89} & \textcolor{BrickRed}{-1.99} & \textcolor{BrickRed}{+1.10} & \textcolor{ForestGreen}{\textbf{+18.62}} & \textcolor{ForestGreen}{\textbf{+14.25}} & \textcolor{BrickRed}{+4.36} & \textcolor{ForestGreen}{\textbf{+6.44}} & \textcolor{ForestGreen}{\textbf{+5.28}} & \textcolor{BrickRed}{+1.16} & \textcolor{BrickRed}{-4.69} & \textcolor{ForestGreen}{\textbf{+9.04}} & \textcolor{ForestGreen}{\textbf{-13.73}} & \textcolor{ForestGreen}{\textbf{+0.46}} & \textcolor{BrickRed}{-8.33} & \textcolor{BrickRed}{+8.79} & \textcolor{BrickRed}{-2.51} & \textcolor{ForestGreen}{\textbf{+18.24}} & \textcolor{ForestGreen}{\textbf{-20.75}} \\
\textit{\method ($c{=}1.00$)} & \textcolor{ForestGreen}{\textbf{+1.27}} & \textcolor{BrickRed}{-1.00} & \textcolor{BrickRed}{+2.27} & \textcolor{ForestGreen}{\textbf{+18.48}} & \textcolor{ForestGreen}{\textbf{+22.55}} & \textcolor{ForestGreen}{\textbf{-4.07}} & \textcolor{ForestGreen}{\textbf{+2.59}} & \textcolor{ForestGreen}{\textbf{+4.29}} & \textcolor{ForestGreen}{\textbf{-1.69}} & \textcolor{BrickRed}{-1.73} & \textcolor{ForestGreen}{\textbf{+16.97}} & \textcolor{ForestGreen}{\textbf{-18.70}} & \textcolor{BrickRed}{-2.45} & \textcolor{BrickRed}{-10.86} & \textcolor{BrickRed}{+8.41} & \textcolor{BrickRed}{-0.80} & \textcolor{ForestGreen}{\textbf{+12.25}} & \textcolor{ForestGreen}{\textbf{-13.04}} \\
\midrule
LoRAM & 11.89 & 10.00 & 1.89 & 21.58 & 5.69 & 15.88 & 10.16 & 8.91 & 1.26 & 27.22 & 24.25 & 2.97 & 24.85 & 31.58 & -6.73 & 24.00 & 12.95 & 11.05 \\
\textit{\method{} ($c{=}0.50$)} & \textcolor{BrickRed}{-2.35} & \textcolor{BrickRed}{-0.96} & \textcolor{ForestGreen}{\textbf{-1.38}} & \textcolor{ForestGreen}{\textbf{+14.13}} & \textcolor{ForestGreen}{\textbf{+26.06}} & \textcolor{ForestGreen}{\textbf{-11.93}} & \textcolor{ForestGreen}{\textbf{+3.09}} & \textcolor{ForestGreen}{\textbf{+4.72}} & \textcolor{ForestGreen}{\textbf{-1.63}} & \textcolor{BrickRed}{-3.40} & \textcolor{BrickRed}{-0.74} & \textcolor{ForestGreen}{\textbf{-2.66}} & \textcolor{ForestGreen}{\textbf{+0.67}} & \textcolor{BrickRed}{-17.20} & \textcolor{BrickRed}{+17.88} & \textcolor{BrickRed}{-1.17} & \textcolor{BrickRed}{-2.95} & \textcolor{BrickRed}{+1.77} \\
\textit{\method{} ($c{=}0.75$)} & \textcolor{BrickRed}{-2.48} & \textcolor{BrickRed}{-0.92} & \textcolor{ForestGreen}{\textbf{-1.56}} & \textcolor{ForestGreen}{\textbf{+15.88}} & \textcolor{ForestGreen}{\textbf{+17.61}} & \textcolor{ForestGreen}{\textbf{-1.73}} & \textcolor{ForestGreen}{\textbf{+6.49}} & \textcolor{ForestGreen}{\textbf{+6.65}} & \textcolor{ForestGreen}{\textbf{-0.16}} & \textcolor{BrickRed}{-4.64} & \textcolor{BrickRed}{-2.68} & \textcolor{ForestGreen}{\textbf{-1.96}} & \textcolor{ForestGreen}{\textbf{+2.11}} & \textcolor{BrickRed}{-14.82} & \textcolor{BrickRed}{+16.92} & \textcolor{BrickRed}{-0.19} & \textcolor{ForestGreen}{\textbf{+9.61}} & \textcolor{ForestGreen}{\textbf{-9.80}} \\
\textit{\method{} ($c{=}1.00$)} & \textcolor{BrickRed}{-0.32} & \textcolor{ForestGreen}{\textbf{+0.07}} & \textcolor{ForestGreen}{\textbf{-0.39}} & \textcolor{ForestGreen}{\textbf{+15.74}} & \textcolor{ForestGreen}{\textbf{+25.91}} & \textcolor{ForestGreen}{\textbf{-10.16}} & \textcolor{ForestGreen}{\textbf{+2.64}} & \textcolor{ForestGreen}{\textbf{+5.65}} & \textcolor{ForestGreen}{\textbf{-3.01}} & \textcolor{BrickRed}{-1.68} & \textcolor{ForestGreen}{\textbf{+5.24}} & \textcolor{ForestGreen}{\textbf{-6.93}} & \textcolor{BrickRed}{-0.80} & \textcolor{BrickRed}{-17.35} & \textcolor{BrickRed}{+16.54} & \textcolor{ForestGreen}{\textbf{+1.53}} & \textcolor{ForestGreen}{\textbf{+3.61}} & \textcolor{ForestGreen}{\textbf{-2.09}} \\
\midrule
LoRA-GA & 11.75 & 10.91 & 0.84 & 36.99 & 32.81 & 4.19 & 14.48 & 14.47 & 0.02 & 22.76 & 17.62 & 5.14 & 24.87 & 16.51 & 8.37 & 21.29 & 14.52 & 6.77 \\
\textit{\method{} ($c{=}0.50$)} & \textcolor{BrickRed}{-2.21} & \textcolor{BrickRed}{-1.88} & \textcolor{ForestGreen}{\textbf{-0.33}} & \textcolor{BrickRed}{-1.29} & \textcolor{BrickRed}{-1.06} & \textcolor{ForestGreen}{\textbf{-0.23}} & \textcolor{BrickRed}{-1.23} & \textcolor{BrickRed}{-0.84} & \textcolor{ForestGreen}{\textbf{-0.39}} & \textcolor{ForestGreen}{\textbf{+1.06}} & \textcolor{ForestGreen}{\textbf{+5.89}} & \textcolor{ForestGreen}{\textbf{-4.83}} & \textcolor{ForestGreen}{\textbf{+0.65}} & \textcolor{BrickRed}{-2.13} & \textcolor{BrickRed}{+2.78} & \textcolor{ForestGreen}{\textbf{+1.54}} & \textcolor{BrickRed}{-4.52} & \textcolor{BrickRed}{+6.05} \\
\textit{\method{} ($c{=}0.75$)} & \textcolor{BrickRed}{-2.35} & \textcolor{BrickRed}{-1.84} & \textcolor{ForestGreen}{\textbf{-0.51}} & \textcolor{ForestGreen}{\textbf{+0.46}} & \textcolor{BrickRed}{-9.51} & \textcolor{BrickRed}{+9.97} & \textcolor{ForestGreen}{\textbf{+2.17}} & \textcolor{ForestGreen}{\textbf{+1.09}} & \textcolor{BrickRed}{+1.08} & \textcolor{BrickRed}{-0.18} & \textcolor{ForestGreen}{\textbf{+3.95}} & \textcolor{ForestGreen}{\textbf{-4.12}} & \textcolor{ForestGreen}{\textbf{+2.08}} & \textcolor{ForestGreen}{\textbf{+0.26}} & \textcolor{BrickRed}{+1.82} & \textcolor{ForestGreen}{\textbf{+2.52}} & \textcolor{ForestGreen}{\textbf{+8.04}} & \textcolor{ForestGreen}{\textbf{-5.52}} \\
\textit{\method{} ($c{=}1.00$)} & \textcolor{BrickRed}{-0.19} & \textcolor{BrickRed}{-0.85} & \textcolor{BrickRed}{+0.67} & \textcolor{ForestGreen}{\textbf{+0.33}} & \textcolor{BrickRed}{-1.21} & \textcolor{BrickRed}{+1.54} & \textcolor{BrickRed}{-1.68} & \textcolor{ForestGreen}{\textbf{+0.10}} & \textcolor{ForestGreen}{\textbf{-1.78}} & \textcolor{ForestGreen}{\textbf{+2.77}} & \textcolor{ForestGreen}{\textbf{+11.87}} & \textcolor{ForestGreen}{\textbf{-9.10}} & \textcolor{BrickRed}{-0.83} & \textcolor{BrickRed}{-2.27} & \textcolor{BrickRed}{+1.44} & \textcolor{ForestGreen}{\textbf{+4.23}} & \textcolor{ForestGreen}{\textbf{+2.04}} & \textcolor{BrickRed}{+2.19} \\
\bottomrule
\end{tabular}
}
\end{table*}

\begin{table*}[!b]
\centering
\caption{\mgp and \mip preservation across all sequences (rank 64). The first row reports absolute scores for Vanilla LoRA; all remaining rows show $\Delta$ relative to Vanilla LoRA. \method variants remain close to the reference on \mgp across all sequences, indicating that the substantial \mfp and forgetting gains in Table~\ref{tab:iclr_delta_all_r64} incur negligible cost to general capability.}
\label{tab:gp_ip_all_r64}
\scriptsize
\setlength{\tabcolsep}{3pt}

\newcommand{\dd}[1]{{\footnotesize$-$#1}}

\resizebox{\linewidth}{!}{%
\begin{tabular}{l|rr|rr|rr|rr|rr|rr}
\toprule
 & \multicolumn{2}{c|}{G1} & \multicolumn{2}{c|}{G2} & \multicolumn{2}{c|}{TRACE} & \multicolumn{2}{c|}{Opp1} & \multicolumn{2}{c|}{Opp2} & \multicolumn{2}{c}{Opp3} \\ \cmidrule(l){2-13}
Method & $\Delta$\mgp & $\Delta$\mip & $\Delta$\mgp & $\Delta$\mip & $\Delta$\mgp & $\Delta$\mip & $\Delta$\mgp & $\Delta$\mip & $\Delta$\mgp & $\Delta$\mip & $\Delta$\mgp & $\Delta$\mip \\
\midrule
Vanilla LoRA \scriptsize{(ref.)} 
  & 51.03 & 58.40 
  & 52.05 & 59.41 
  & 54.59 & 60.39 
  & 52.41 & 60.26 
  & 55.63 & 61.43 
  & 52.47 & 60.09 \\
\midrule
LoRAM
  & \dd{0.10} & \dd{1.86}
  & \dd{0.78} & \dd{5.34}
  & \dd{2.33} & \dd{3.14}
  & \dd{1.88} & \dd{4.84}
  & \dd{1.84} & \dd{0.62}
  & \dd{1.49} & \dd{2.57} \\
LoRA-GA
  & 0.00 & \dd{10.11}
  & \dd{5.18} & \dd{20.28}
  & \dd{1.48} & \dd{4.96}
  & \dd{2.34} & \dd{8.52}
  & \dd{2.05} & \dd{2.17}
  & \dd{2.58} & \dd{6.63} \\
\midrule
\method{} ($c{=}0.50$)
  & \dd{0.10} & \dd{5.61}
  & \dd{4.38} & \dd{21.08}
  & \dd{2.24} & \dd{2.72}
  & \dd{2.27} & \dd{6.61}
  & \dd{2.74} & \dd{1.80}
  & \dd{1.51} & \dd{7.30} \\
\method{} ($c{=}0.75$)
  & \dd{0.42} & \dd{3.00}
  & \dd{4.78} & \dd{22.37}
  & \dd{1.53} & \dd{4.66}
  & \dd{1.64} & \dd{5.27}
  & \dd{2.24} & \dd{5.07}
  & \dd{2.14} & \dd{3.95} \\
\method{} ($c{=}1.00$)
  & \dd{0.04} & \dd{3.94}
  & \dd{4.66} & \dd{13.18}
  & \dd{1.76} & \dd{2.39}
  & \dd{2.12} & \dd{8.65}
  & \dd{2.80} & \dd{4.33}
  & \dd{1.67} & \dd{6.61} \\
\bottomrule
\end{tabular}
}
\end{table*}

\subsection{Results} 
\label{sec:results}

Table~\ref{tab:iclr_delta_all_r64} displays that \method consistently improves \mfp and reduces \mfgt across task sequences in comparison with baselines, with the most pronounced gains on sequences characterized by severe catastrophic forgetting. 
When compared to Vanilla LoRA, \method ($c=1.0$) achieves \mfp improvements of up to +22.55 on G2 and +16.97 on Opp1, while simultaneously reducing \mfgt by up to 18.70 and 20.75 points on Opp1 and Opp3, respectively — sequences where the baseline exhibits the highest inter-task interference. 
Similar trends hold for LoRAM, where \method ($c=0.50$) recovers up to +26.06 \mfp points on G2 and reduces \mfgt by 11.93. 
Gains on \map are more modest, consistent with the fact that \method targets the retention of previously acquired knowledge rather than peak task-specific performance during training. 
The comparatively smaller gains observed with LoRA-GA indicate that its initialization already exhibits favorable properties that partially address catastrophic forgetting — yet our approach further enhances these gains beyond what LoRA-GA alone achieves. These trends persist at rank $r{=}128$, as reported in Appendix~\ref{app:rank128exp}.

\begin{figure}[!t]
    \centering
    \includegraphics[width=\textwidth]{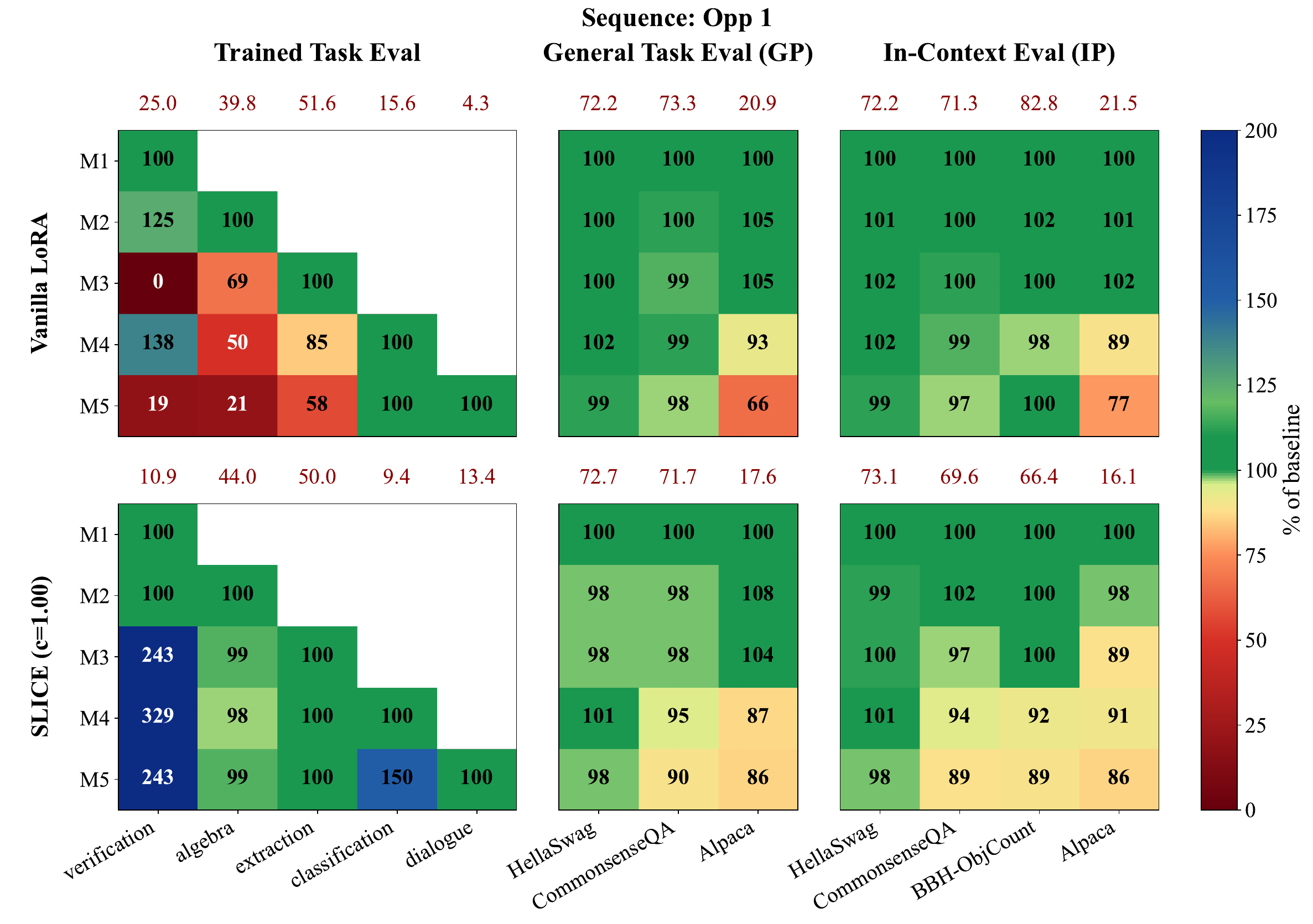}
    \caption{\textbf{Performance heatmaps comparing Vanilla LoRA vs.\ \method ($c=1.0$)\ during CL on \textsc{NI-Seq-Opposite-1}.}
    Numbers above each heatmap indicate baseline performance on the trained task, \mgp and \mip, at left, center, and right, respectively. 
    Heatmap values show percentage change relative to baseline.}
    \label{fig:opp1-comparison}
\end{figure}

Table~\ref{tab:gp_ip_all_r64} shows that \method variants incur modest reductions in \mgp and \mip relative to Vanilla LoRA across most sequences, remaining close to the reference across all evaluated sequences. The exception is \mip on \textsc{NI-Seq-G2}, where larger drops are observed. 
Crucially, however, these reductions in general capability occur alongside substantially larger gains in task retention, indicating that the gradient surgery applied by \method primarily reallocates optimization pressure toward mitigating catastrophic forgetting rather than compressing the model's underlying representational capacity. 
This asymmetry in the trade-off---small \mgp cost against large \mfp and \mfgt gains---suggests that \method reduces destructive gradient interference between sequential tasks without significantly degrading broad linguistic knowledge. 
Moreover, the $\alpha$ sweep in Appendix~\ref{app:alpha-sweep} demonstrates that the \mgp and \mip cost is largely recoverable: reducing $\alpha$ from 2 to 1 closes much of the \mip gap on G2 while leaving \map, \mfp, and \mfgt nearly unchanged, indicating that $\alpha$ serves as a reliable control knob for the generalization--retention trade-off without requiring a different initialization strategy.

Figure~\ref{fig:opp1-comparison} shows the stability–plasticity dynamics on \textsc{NI-Seq-Opposite-1}, comparing Vanilla LoRA (highest \map baseline) with \method ($c=1.0$).
Vanilla LoRA exhibits strong performance across all tasks at each step, as indicated by the top values in the left heatmap.
However, its off-diagonal entries decay rapidly as subsequent tasks are learned: by the final stage M5, tasks 1 and 2 have collapsed to 19\% and 21\% of the baseline score, revealing severe catastrophic forgetting. 
In contrast, \method ($c=1.0$) maintains substantially higher off-diagonal retention throughout the sequence: at M5, task 1 not only avoids collapse but reaches 243\% of baseline (26.5 points, higher than vanilla baseline) while tasks 3–4 remain similar. 
This pattern confirms that \method achieves a competitive \map while delivering markedly higher \mfp and lower \mfgt, precisely because the conflict-aware initialization steers the adapter subspace away from directions that would overwrite earlier task knowledge. 

\method shows a mild reduction relative to Vanilla LoRA on \mgp and \mip. 
Nevertheless, these marginal losses in general capability are small in comparison to the CL gains, as shown in Table~\ref{tab:iclr_delta_all_r64}. 
Opp1 shows an \mfp improvement of 16.97 and a \mfgt reduction of 18.70 points, indicating that the gradient surgery reallocates representational capacity toward task retention without meaningfully eroding the model's broad linguistic competence.
Appendix \ref{app:heatmaps} extends this comparison to LoRA-GA and LoRAM. 
\method remains the only approach that does not degenerate \mfp for any task (including Algebra, where all other methods perform poorly) while remaining competitive on \mgp and \mip.

Our results are robust to a known confounder in LoRA initialization comparisons. Zhang et al. \citep{zhang2025primacy} showed that spectral initialization methods can appear beneficial due to implicit magnitude amplification rather than directional quality. 
Our experimental protocol applies variance-matched magnitude rescaling (\hyperref[stg:mag_rescaling]{Stage~4}) across all initialization baselines—\method, LoRA-GA, and LoRAM—so the gains reported reflect genuine subspace selection rather than scale artifacts.

\section{Related Work}


\paragraph{Initialization Strategies for LoRA.} The standard LoRA initialization scheme~\citep{hu2022lora}, with Gaussian noise for $A$ and zeros for $B$, can limit convergence speed and downstream performance, motivating alternative initialization strategies. 
Existing methods can be broadly grouped into spectral approaches, which derive adapter parameters from pre-trained weights, and gradient-based approaches, which use task-specific gradient information. 
Spectral methods such as PiSSA~\citep{meng2024pissa} and MiLoRA~\citep{wang2024milora} initialize adapters from principal or minor components of the pre-trained weights. 
Zhang et al.~\citep{zhang2025primacy} argued that much of the benefit of these methods may arise from increased update magnitude rather than from the specific spectral directions, and proposed LoRAM as a simpler alternative based on deterministic orthogonal bases constructed with the Discrete Sine Transform. 
In contrast, LoRA-GA~\citep{wang2024loraga} initializes adapters by computing gradients on a small calibration dataset and decomposing the resulting gradient matrices to obtain a task-relevant subspace. 

Based on these findings, we selected LoRAM and LoRA-GA as initialization baselines representing magnitude-oriented and gradient-based initialization strategies, respectively. 
\method extends LoRA-GA: while LoRA-GA initializes adapters using only the adaptation task gradient $G_{\text{cur}}$, \method also incorporates the preserve task gradient $G_{prev}$ and projects out conflicting components before decomposition.

\paragraph{LoRA-Based Continual Learning.} A growing body of work leverages low-rank adapters to mitigate catastrophic forgetting in CL, predominantly through orthogonality constraints imposed \emph{during training}. O-LoRA~\citep{wang2023olora} learns each task within a low-rank subspace kept orthogonal to the subspaces of all previous tasks. InfLoRA~\citep{liang2024inflora} designs the dimensionality reduction matrix $B_t$ such that the update subspace is orthogonal to the gradient subspace of previously learned tasks. OPLoRA~\citep{xiong2026oplora} constrains updates to the orthogonal complement of the dominant singular subspace of the pre-trained weights. 
Beyond orthogonality, AM-LoRA~\citep{liu2024learningattentionalmixtureloras} introduces an attentional mixture of task-specific LoRA modules with learned routing. 
Crucially, all of these methods address the stability-plasticity trade-off through constraints or architectural mechanisms applied \emph{throughout the training process}, while remaining agnostic to how the adapter parameters are initialized. 
\method is not a competitor to these methods but rather a complementary component that occupies a different stage of the pipeline: it determines the initial values of $A$ and $B$ \emph{before any training begins}, and the resulting adapters can then be trained with any of the above strategies. 

\paragraph{Gradient surgery in multi-task optimization.}
Multi-task optimization (MTO) replaces the naive average of per-task gradients with a corrected update direction that mitigates destructive interference. PCGrad~\citep{yu2020gradient_surgery} projects each task gradient onto the normal plane of any conflicting partner whenever their cosine similarity is negative. GradVac~\citep{wang2021gradvac} generalizes this surgery by aligning gradients to a target cosine similarity tracked online with an exponential moving average rather than to strict orthogonality, recovering PCGrad as the special case of a zero target. 
MGDA~\citep{sener2018mgda}, which itself returns the minimum-norm convex combination of task gradients and converges to a Pareto-stationary point. GradDrop~\citep{chen2020graddrop} stochastically masks coordinates whose task gradients disagree in sign, biasing the optimizer toward joint minima rather than task-specific ones. Aligned-MTL~\citep{senushkin2023alignedmtl} stabilizes training by enforcing a unit condition number on the gradient matrix via a Procrustes-style SVD, simultaneously eliminating directional conflict and magnitude dominance. Nash-MTL~\citep{navon2022nashmtl} casts the gradient-combination step as a cooperative bargaining game and solves for the weights realizing the Nash equilibrium over directional utilities, yielding a scale-invariant aggregator. RotoGrad~\citep{javaloy2022rotograd} pairs magnitude rebalancing with learned task-specific rotations of the shared feature space to homogenize both the size and the direction of gradient updates. FAMO~\citep{liu2023famo} amortizes the cost of gradient surgery to $\mathcal{O}(1)$ per step by maintaining softmax-parameterized task weights updated from temporal differences in scalar log-losses.


\section{Conclusion}
\label{sec:conclusion}

We introduced \method, a LoRA initialization method based on gradient surgery that reconciles the trade-off between learning new tasks while maintaining the capabilities of previously learned tasks. 
\method substantially improves stability without a proportionate decrease in plasticity. 

In sequences exhibiting severe inter-task gradient conflict, particularly the NI-SEQ-OPPOSITE, \method yields \mfp gains of up to +22.55 points and \mfgt reductions exceeding 20 points over vanilla LoRA, while \map remains competitive. 
\method is composable by design: since it operates exclusively at initialization, the resulting adapter pairs serve as drop-in replacements compatible with any downstream training-time continual learning strategy.

\method has a potentially negative impact on \mgp and \mip, with mild \mgp reductions. 
While this effect is small relative to CL improvements, they indicate that the projected gradient subspace, although conflict-free, may sacrifice some alignment with directions beneficial to broad linguistic generalization. 
A selection of the $\alpha$ hyperparameter can diminish this effect.

Our evaluation is designed to isolate the effect of subspace selection from confounding factors, such as implicit magnitude amplification, by controlling for them via variance-matched rescaling across all baselines. We further introduce NI-SEQ-OPPOSITE, an adversarial sequences of maximally gradient-conflicting task sequences, complementing standard TRACE, G1 and G2 evaluations to provide a complete picture of each method's behavior across the full spectrum of inter-task interference.

We show that the initialization of low-rank adapters is a consequential and underexplored intervention point for CL: a single, lightweight gradient-surgery step prior to training yields stability gains superior to other initialization methods. 
Future work can expand on conflict-aware subspace selection at initialization beyond the projection operators studied here and use our conflicting task sequence NI-SEQ-OPPOSITE to evaluate the continual adaptation of LLMs.

\subsection{Limitations}
\label{sec:limitations}

\paragraph{Computational cost considerations.} While LoRA adapters reduce the parameter count requiring gradient updates and thus improve training efficiency relative to full fine-tuning, \method' gradient-based initialization incurs non-trivial computational overhead during setup. \method includes backward passes over batches of $\mathcal{D}_{cur}$ and ${\mathcal{D}}_{prev}$. Additionally, the cost of the randomized SVD scales as $\mathcal{O}(\min(m^2 n, m n^2))$ per target matrix for dimensions $m, n$. For practitioners operating under strict constraints, this overhead may still be prohibitive relative to LoRAM's essentially free deterministic initialization.

\subsection{Broader Impact}
\label{sec:broader-impact}

Continual fine-tuning of large language models is increasingly common in deployed systems, where models must adapt sequentially to new domains, instructions, and user requirements. 
Methods that reduce catastrophic forgetting in this regime lower the practical barrier to maintaining capable models over time, without the cost of full retraining. 
By providing a lightweight, composable initialization step that improves task retention, \method helps make sequential adaptation more resource-efficient.
This is relevant to both research labs and practitioners operating under resource constraints.

We also introduce \textsc{NI-Seq-Opposite}, a set of sequences for evaluating continual learning methods under adversarial gradient conflict.
This resource can help future work measure robustness to inter-task interference in a controlled, reproducible way, independent of the proposed method.




\subsubsection*{Acknowledgments}
\label{chap:ack}
This study was financed in part by the Coordination for the Improvement of Higher Education Personnel (CAPES) --- Finance Code 001; by Conselho Nacional de Desenvolvimento Científico e Tecnológico (CNPq)--- Grant Number: 443072/2024-8; and by Fundação de Amparo à Pesquisa do Estado do Rio Grande do Sul (FAPERGS) --- Grant Number: 25/2551-0000891-3.

This work was supported by Kunumi Institute. The authors thank the institution for its financial support and commitment to advancing scientific research.

{
\small

\bibliography{refs}

@inproceedings{wang-etal-2022-super,
    title = "Super-{N}atural{I}nstructions: Generalization via Declarative Instructions on 1600+ {NLP} Tasks",
    author = "Wang, Yizhong  and
      Mishra, Swaroop  and
      Alipoormolabashi, Pegah  and
      Kordi, Yeganeh  and
      Mirzaei, Amirreza  and
      Naik, Atharva  and
      Ashok, Arjun  and
      Dhanasekaran, Arut Selvan  and
      Arunkumar, Anjana  and
      Stap, David  and
      Pathak, Eshaan  and
      Karamanolakis, Giannis  and
      Lai, Haizhi  and
      Purohit, Ishan  and
      Mondal, Ishani  and
      Anderson, Jacob  and
      Kuznia, Kirby  and
      Doshi, Krima  and
      Pal, Kuntal Kumar  and
      Patel, Maitreya  and
      Moradshahi, Mehrad  and
      Parmar, Mihir  and
      Purohit, Mirali  and
      Varshney, Neeraj  and
      Kaza, Phani Rohitha  and
      Verma, Pulkit  and
      Puri, Ravsehaj Singh  and
      Karia, Rushang  and
      Doshi, Savan  and
      Sampat, Shailaja Keyur  and
      Mishra, Siddhartha  and
      Reddy A, Sujan  and
      Patro, Sumanta  and
      Dixit, Tanay  and
      Shen, Xudong",
    editor = "Goldberg, Yoav  and
      Kozareva, Zornitsa  and
      Zhang, Yue",
    booktitle = "Proceedings of the 2022 Conference on Empirical Methods in Natural Language Processing",
    month = dec,
    year = "2022",
    address = "Abu Dhabi, United Arab Emirates",
    publisher = "Association for Computational Linguistics",
    url = "https://aclanthology.org/2022.emnlp-main.340/",
    doi = "10.18653/v1/2022.emnlp-main.340",
    pages = "5085--5109"
}

@misc{wang2023tracecomprehensivebenchmarkcontinual,
      title={TRACE: A Comprehensive Benchmark for Continual Learning in Large Language Models}, 
      author={Xiao Wang and Yuansen Zhang and Tianze Chen and Songyang Gao and Senjie Jin and Xianjun Yang and Zhiheng Xi and Rui Zheng and Yicheng Zou and Tao Gui and Qi Zhang and Xuanjing Huang},
      year={2023},
      eprint={2310.06762},
      archivePrefix={arXiv},
      primaryClass={cs.CL},
      url={https://arxiv.org/abs/2310.06762}, 
}

@article{shi2025continualsurvey,
author = {Shi, Haizhou and Xu, Zihao and Wang, Hengyi and Qin, Weiyi and Wang, Wenyuan and Wang, Yibin and Wang, Zifeng and Ebrahimi, Sayna and Wang, Hao},
title = {Continual Learning of Large Language Models: A Comprehensive Survey},
year = {2025},
issue_date = {April 2026},
publisher = {Association for Computing Machinery},
address = {New York, NY, USA},
volume = {58},
number = {5},
issn = {0360-0300},
url = {https://doi.org/10.1145/3735633},
doi = {10.1145/3735633},
journal = {ACM Comput. Surv.},
month = nov,
articleno = {120},
numpages = {42}
}

@article{
kirkpatrick2017overcoming,
author = {James Kirkpatrick  and Razvan Pascanu  and Neil Rabinowitz  and Joel Veness  and Guillaume Desjardins  and Andrei A. Rusu  and Kieran Milan  and John Quan  and Tiago Ramalho  and Agnieszka Grabska-Barwinska  and Demis Hassabis  and Claudia Clopath  and Dharshan Kumaran  and Raia Hadsell },
title = {Overcoming catastrophic forgetting in neural networks},
journal = {Proceedings of the National Academy of Sciences},
volume = {114},
number = {13},
pages = {3521-3526},
year = {2017},
doi = {10.1073/pnas.1611835114},
URL = {https://www.pnas.org/doi/abs/10.1073/pnas.1611835114},
eprint = {https://www.pnas.org/doi/pdf/10.1073/pnas.1611835114}}

@inproceedings{zhang2025primacy,
title={The Primacy of Magnitude in Low-Rank Adaptation},
author={Zicheng Zhang and Haoran Li and Yifeng Zhang and Guoqiang Gong and Jiaxing Wang and Junxing Hu and pengzhang liu and Qixia Jiang},
booktitle={The Thirty-ninth Annual Conference on Neural Information Processing Systems},
year={2026},
url={https://openreview.net/forum?id=s4LnWgjacg}
}

@inproceedings{meng2024pissa,
  title={{PiSSA}: Principal Singular Values and Singular Vectors Adaptation of Large Language Models},
  author={Meng, Fanxu and Wang, Zhaohui and Zhang, Muhan},
  booktitle={Advances in Neural Information Processing Systems (NeurIPS)},
  year={2024},
  url={https://openreview.net/forum?id=6ZBHIEtdP4}
}

@inproceedings{wang2024milora,
  title={MiLoRA: Harnessing Minor Singular Components for Parameter-Efficient LLM Finetuning},
  author={Wang, Hanqing and Xiao, Zeguan and Li, Yixia and Wang, Shuo and Chen, Guanhua and Chen, Yun},
  booktitle={Annual Conference of the North American Chapter of the Association for Computational Linguistics (NAACL)},
  year={2025},
  url ={https://aclanthology.org/2025.naacl-long.248/}
}

@inproceedings{wang2024loraga,
  title={{LoRA-GA}: Low-Rank Adaptation with Gradient Approximation},
  author={Wang, Shaowen and Yu, Linxi and Li, Jian},
  booktitle={Advances in Neural Information Processing Systems (NeurIPS)},
  year={2024},
  url={https://openreview.net/forum?id=VaLAWrLHJv}
}

@inproceedings{hu2022lora,
  title={{LoRA}: Low-Rank Adaptation of Large Language Models},
  author={Hu, Edward J. and Shen, Yelong and Wallis, Phillip and Allen-Zhu, Zeyuan and Li, Yuanzhi and Wang, Shean and Wang, Lu and Chen, Weizhu},
  booktitle={International Conference on Learning Representations (ICLR)},
  year={2022},
  url={https://openreview.net/forum?id=nZeVKeeFYf9}
}

@inproceedings{wang2023olora,
    title = "Orthogonal Subspace Learning for Language Model Continual Learning",
    author = "Wang, Xiao  and
      Chen, Tianze  and
      Ge, Qiming  and
      Xia, Han  and
      Bao, Rong  and
      Zheng, Rui  and
      Zhang, Qi  and
      Gui, Tao  and
      Huang, Xuanjing",
    editor = "Bouamor, Houda  and
      Pino, Juan  and
      Bali, Kalika",
    booktitle = "Findings of the Association for Computational Linguistics: EMNLP 2023",
    month = dec,
    year = "2023",
    address = "Singapore",
    publisher = "Association for Computational Linguistics",
    url = "https://aclanthology.org/2023.findings-emnlp.715/",
    doi = "10.18653/v1/2023.findings-emnlp.715",
    pages = "10658--10671"
}

@article{xiong2026oplora, 
title={OPLoRA: Orthogonal Projection LoRA Prevents Catastrophic Forgetting During Parameter-Efficient Fine-Tuning}, 
volume={40}, 
url={https://ojs.aaai.org/index.php/AAAI/article/view/40703}, 
DOI={10.1609/aaai.v40i40.40703}, 
abstractNote={Low-Rank Adaptation (LoRA) enables efficient fine-tuning of large language models but suffers from catastrophic forgetting when learned updates interfere with the dominant singular directions that encode essential pre-trained knowledge. We propose Orthogonal Projection LoRA (OPLoRA), a theoretically grounded approach that prevents this interference through double-sided orthogonal projections. By decomposing frozen weights via SVD, OPLoRA constrains LoRA updates to lie entirely within the orthogonal complement of the top-k singular subspace using projections PL = I − Uk Ukᵀ and PR = I − Vk Vkᵀ. We prove that this construction exactly preserves the top-k singular triples, providing mathematical guarantees for knowledge retention. To quantify subspace interference, we introduce ρk, a metric measuring update alignment with dominant directions. Extensive experiments across commonsense reasoning, mathematics, and code generation demonstrate that OPLoRA significantly reduces forgetting while maintaining competitive task-specific performance on LLaMA-2 7B and Qwen2.5 7B, establishing orthogonal projection as an effective mechanism for knowledge preservation in parameter-efficient fine-tuning.}, 
number={40}, 
journal={Proceedings of the AAAI Conference on Artificial Intelligence}, 
author={Xiong, Yifeng and Xie, Xiaohui}, 
year={2026}, 
month={Mar.}, 
pages={34088-34096} 
}

@inproceedings{liang2024inflora,
  title={InfLoRA: Interference-Free Low-Rank Adaptation for Continual Learning},
  author={Liang, Yan-Shuo and Li, Wu-Jun},
  booktitle={Proceedings of the IEEE/CVF Conference on Computer Vision and Pattern Recognition},
  pages={23638--23647},
  year={2024}
}

@misc{liu2024learningattentionalmixtureloras,
      title={Learning Attentional Mixture of LoRAs for Language Model Continual Learning}, 
      author={Jialin Liu and Jianhua Wu and Jie Liu and Yutai Duan},
      year={2024},
      eprint={2409.19611},
      archivePrefix={arXiv},
      primaryClass={cs.CL},
      url={https://arxiv.org/abs/2409.19611}, 
}

@misc{eval-harness,
  author       = {Gao, Leo and Tow, Jonathan and Abbasi, Baber and Biderman, Stella and Black, Sid and DiPofi, Anthony and Foster, Charles and Golding, Laurence and Hsu, Jeffrey and Le Noac'h, Alain and Li, Haonan and McDonell, Kyle and Muennighoff, Niklas and Ociepa, Chris and Phang, Jason and Reynolds, Laria and Schoelkopf, Hailey and Skowron, Aviya and Sutawika, Lintang and Tang, Eric and Thite, Anish and Wang, Ben and Wang, Kevin and Zou, Andy},
  title        = {The Language Model Evaluation Harness},
  month        = 07,
  year         = 2024,
  publisher    = {Zenodo},
  version      = {v0.4.3},
  doi          = {10.5281/zenodo.12608602},
  url          = {https://zenodo.org/records/12608602}
}

@inproceedings{
wang2021gradvac,
title={Gradient Vaccine: Investigating and Improving Multi-task Optimization in Massively Multilingual Models},
author={Zirui Wang and Yulia Tsvetkov and Orhan Firat and Yuan Cao},
booktitle={International Conference on Learning Representations},
year={2021},
url={https://openreview.net/forum?id=F1vEjWK-lH_}
}

@article{chen2020graddrop,
  title={Just pick a sign: Optimizing deep multitask models with gradient sign dropout},
  author={Chen, Zhao and Ngiam, Jiquan and Huang, Yanping and Luong, Thang and Kretzschmar, Henrik and Chai, Yuning and Anguelov, Dragomir},
  journal={Advances in Neural Information Processing Systems},
  volume={33},
  pages={2039--2050},
  year={2020}
}

@article{sener2018mgda,
  title={Multi-task learning as multi-objective optimization},
  author={Sener, Ozan and Koltun, Vladlen},
  journal={Advances in neural information processing systems},
  volume={31},
  year={2018}
}

@inproceedings{senushkin2023alignedmtl,
  title={Independent component alignment for multi-task learning},
  author={Senushkin, Dmitry and Patakin, Nikolay and Kuznetsov, Arseny and Konushin, Anton},
  booktitle={Proceedings of the IEEE/CVF Conference on Computer Vision and Pattern Recognition},
  pages={20083--20093},
  year={2023}
}

@inproceedings{navon2022nashmtl,
  title={Multi-Task Learning as a Bargaining Game},
  author={Navon, Aviv and Shamsian, Aviv and Achituve, Idan and Maron, Haggai and Kawaguchi, Kenji and Chechik, Gal and Fetaya, Ethan},
  booktitle={International Conference on Machine Learning},
  pages={16428--16446},
  year={2022},
  organization={PMLR}
}

@inproceedings{javaloy2022rotograd,
title={RotoGrad: Gradient Homogenization in Multitask Learning},
author={Adri{\'a}n Javaloy and Isabel Valera},
booktitle={International Conference on Learning Representations},
year={2022},
url={https://openreview.net/forum?id=T8wHz4rnuGL}
}

@article{liu2023famo,
  title={Famo: Fast adaptive multitask optimization},
  author={Liu, Bo and Feng, Yihao and Stone, Peter and Liu, Qiang},
  journal={Advances in Neural Information Processing Systems},
  volume={36},
  pages={57226--57243},
  year={2023}
}

@inproceedings{yu2020gradient_surgery,
 author = {Yu, Tianhe and Kumar, Saurabh and Gupta, Abhishek and Levine, Sergey and Hausman, Karol and Finn, Chelsea},
 booktitle = {Advances in Neural Information Processing Systems},
 editor = {H. Larochelle and M. Ranzato and R. Hadsell and M.F. Balcan and H. Lin},
 pages = {5824--5836},
 publisher = {Curran Associates, Inc.},
 title = {Gradient Surgery for Multi-Task Learning},
 url = {https://proceedings.neurips.cc/paper_files/paper/2020/file/3fe78a8acf5fda99de95303940a2420c-Paper.pdf},
 volume = {33},
 year = {2020}
}

@inproceedings{jiang2025unlocking,
title={Unlocking the Power of Function Vectors for Characterizing and Mitigating Catastrophic Forgetting in Continual Instruction Tuning},
author={Gangwei Jiang and Caigao JIANG and Zhaoyi Li and Siqiao Xue and JUN ZHOU and Linqi Song and Defu Lian and Ying Wei},
booktitle={The Thirteenth International Conference on Learning Representations},
year={2025},
url={https://openreview.net/forum?id=gc8QAQfXv6}
}

@misc{chaudhry2019er,
      title={On Tiny Episodic Memories in Continual Learning}, 
      author={Arslan Chaudhry and Marcus Rohrbach and Mohamed Elhoseiny and Thalaiyasingam Ajanthan and Puneet K. Dokania and Philip H. S. Torr and Marc'Aurelio Ranzato},
      year={2019},
      eprint={1902.10486},
      archivePrefix={arXiv},
      primaryClass={cs.LG},
      url={https://arxiv.org/abs/1902.10486}, 
}

@inproceedings{buzzega2020der,
  title={Dark experience for general continual learning: a strong, simple baseline},
  author={Buzzega, Pietro and Boschini, Matteo and Porrello, Angelo and Abati, Davide and Calderara, Simone},
  booktitle={Advances in Neural Information Processing Systems 33 (NeurIPS 2020)},
  pages={15920--15930},
  year={2020}
}

@inproceedings{chaudhry2018agem,
  title={Efficient Lifelong Learning with A-GEM},
  author={Chaudhry, Arslan and Ranzato, Marc’Aurelio and Rohrbach, Marcus and Elhoseiny, Mohamed},
  booktitle={International Conference on Learning Representations},
  year={2018}
}
}
\appendix


\newpage

\section{Alpha Variation}
\label{app:alpha-sweep}

We sweep rs-LoRA's $\alpha$ for \method ($c{=}0.50$) across three sequences, NI-Seq-Opp3, NI-Seq-G2, and TRACE, to quantify how scaling the adapter update trade-offs generalization (\mgp and \mip) against task metrics (\map, \mfp and \mfgt). 
For each sequence, we report results at $\alpha \in \{1,2,4\}$ and highlight $\Delta(2-1)$ to isolate the effect of a modest reduction in $\alpha$. 


Table~\ref{tab:slice-cagrad050-alpha} shows that decreasing $\alpha$ from 2 to 1, improves \mgp and \mip for NI-Seq-G2 (\mgp: $-4.2$, \mip: $-19.5$) while \map and \mfp change modestly and \mfgt stays small.
The same trend appears in the other sequences, suggesting $\alpha$ is a reliable \mgp and \mip control knob without affecting \map, \mfp and \mfgt. 

\begin{figure*}[!b]
\centering
\begin{subfigure}[t]{0.48\linewidth}
  \centering
  \includegraphics[width=\linewidth]{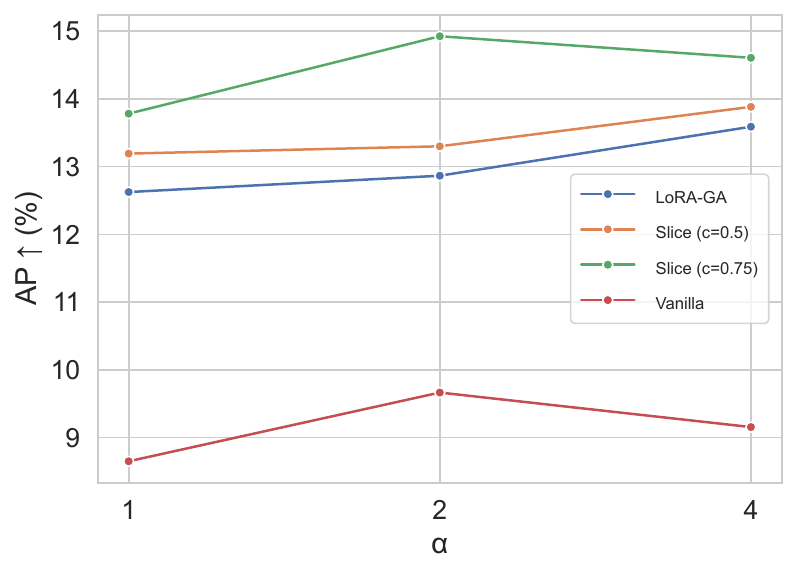}
  \caption{AP}
\end{subfigure}
\begin{subfigure}[t]{0.48\linewidth}
  \centering
  \includegraphics[width=\linewidth]{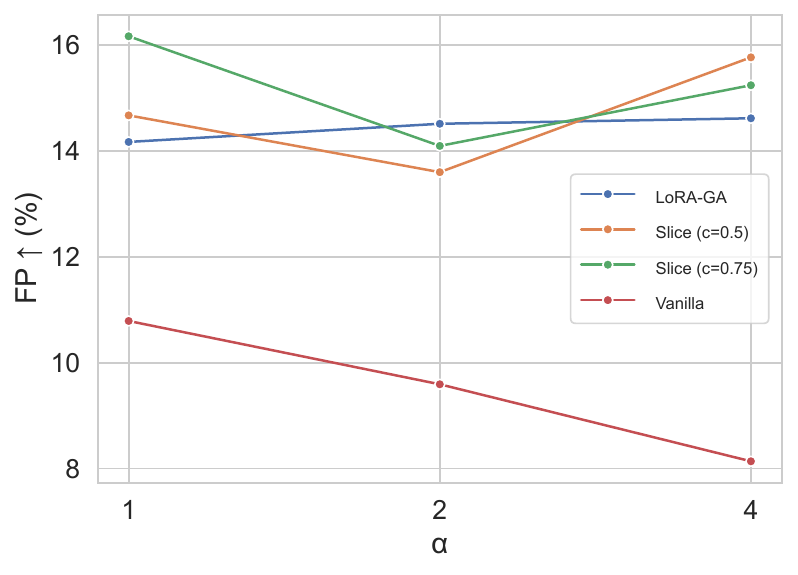}
  \caption{FP}
\end{subfigure}
\begin{subfigure}[t]{0.48\linewidth}
  \centering
  \includegraphics[width=\linewidth]{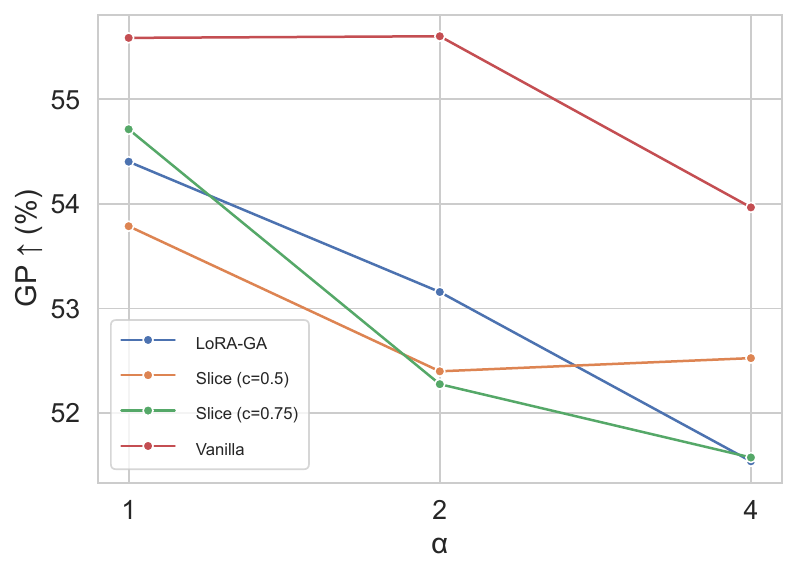}
  \caption{GP}
\end{subfigure}
\begin{subfigure}[t]{0.48\linewidth}
  \centering
  \includegraphics[width=\linewidth]{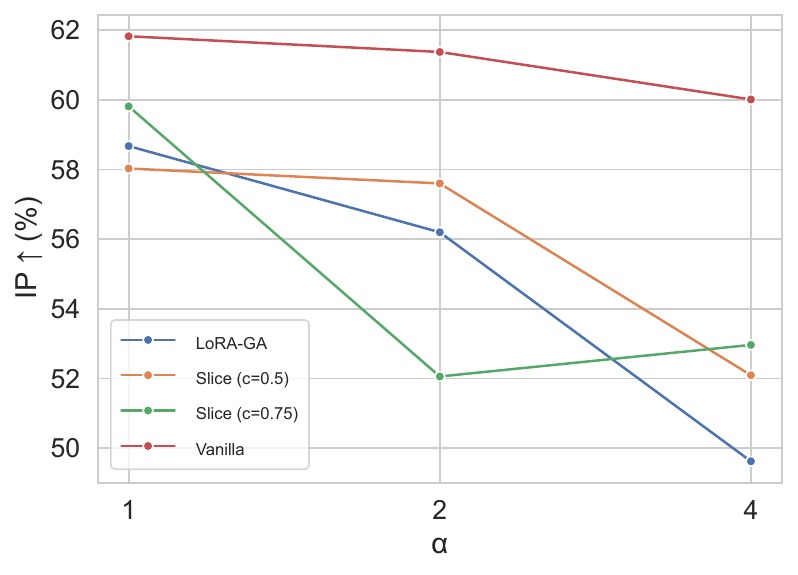}
  \caption{IP}
\end{subfigure}

\caption{Metrics for TRACE across $\alpha \in \{1, 2, 4\}$ for (\method, $c{=}0.50$).}
\label{fig:alpha-sweep-trace}
\end{figure*}

Figure~\ref{fig:alpha-sweep-trace} shows per-metric $\alpha$ on TRACE: \mgp and \mip decrease as $\alpha$ is reduced, while \map, \mfp and \mfgt remain comparatively stable. 
We match the initial adapter variance to the layer weights for LoRA-GA, LoRAM, and \method, but not for vanilla; vanilla LoRA uses a lower initial variance as discussed in \cite{zhang2025primacy}.

\begin{table}[!b]
\centering
\caption{Slice ($c{=}0.50$): effect of $\alpha$ across three sequences. Deltas highlight that $\alpha$ can be used to control \mgp and \mip.}
\label{tab:slice-cagrad050-alpha}
\small
\setlength{\tabcolsep}{6pt}
\renewcommand{\arraystretch}{1.15}
\sisetup{round-mode=places,round-precision=1,detect-weight=true,table-number-alignment=center}
\begin{tabular}{ll S[table-format=+3.1] S[table-format=+3.1] S[table-format=+3.1] S[table-format=+3.1] S[table-format=+3.1]}
\toprule
Sequence & $\alpha$ & {\map} & {\mfp} & {\mgp} & {\mip} & {\mfgt} \\
\midrule
\multirow{5}{*}{NI-Seq-Opp3} 
 & 1 & 25.3 & 26.1 & 53.5 & 59.5 & -0.9 \\
 & 2 & 22.8 & 10.0 & 50.9 & 52.8 & 12.8 \\
 & 4 & 13.1 & 5.9 & 45.9 & 49.4 & 7.1 \\
 & $\Delta(2-1)$ & -2.5 & -16.1 & -2.6 & -6.7 & 13.7 \\
\addlinespace
\multirow{5}{*}{NI-Seq-G2} 
 & 1 & 37.6 & 33.2 & 51.9 & 57.8 & 4.4 \\
 & 2 & 35.7 & 31.7 & 47.7 & 38.3 & 0.0395 \\
 & 4 & 25.4 & 18.0 & 44.4 & 34.3 & 7.4 \\
 & $\Delta(2-1)$ & -1.9 & -1.5 & -4.2 & -19.5 & -4.4 \\
\addlinespace
\multirow{5}{*}{TRACE} 
 & 1 & 13.2 & 14.7 & 53.8 & 58.0 & -1.5 \\
 & 2 & 13.3 & 13.6 & 52.4 & 57.6 & -0.0037 \\
 & 4 & 13.9 & 15.8 & 52.5 & 52.1 & -1.9 \\
 & $\Delta(2-1)$ & 0.1 & -1.1 & -1.4 & -0.4 & 1.5 \\
\bottomrule
\end{tabular}
\end{table}

\subsection{Weight absorption}
\label{app:weight-absorption}

After computing $(A,B)$ for each target linear layer, we write the tensors into the corresponding LoRA adapter parameters \emph{before} training. To preserve the function of the pretrained model at initialization (i.e., ensure the \emph{effective} weight remains $W_0$ even after attaching LoRA), we perform \emph{weight absorption}:
\begin{equation}
W_{\text{eff}} \;=\; W_{\text{base}} + \frac{\alpha}{r}BA,
\end{equation}
we set the frozen base weight to
\begin{equation}
W_{\text{base}} \;\leftarrow\; W_0 - \frac{\alpha}{r}BA,
\label{eq:weight_absorption}
\end{equation}
so that immediately after initialization,
\begin{equation}
W_{\text{eff}} \;=\; \Big(W_0 - \frac{\alpha}{r}BA\Big) + \frac{\alpha}{r}BA \;=\; W_0.
\end{equation}
This absorption step ensures that any observed effect at step $0$ is attributable to training dynamics rather than an unintended initial perturbation of the forward pass.

\section{Experimental Configuration}
\label{app:experimental}

\paragraph{Base model and tokenizer.} All experiments use Llama-3.2-3B-Instruct, trained using PEFT and \texttt{bfloat16} quantization.

\paragraph{Data and task sequences.} Continual learning sequences come from SuperNI and TRACE. SuperNI inputs are formatted as a chat prompt with the task definition and input; TRACE prompts include the task name and instruction. For each task, we shuffle with seed 42 and hold out up to 200 examples for validation; the remaining examples are used for training. 

\paragraph{LoRA adapters.} We use rs-LoRA with rank $r \in \{64,128\}$ (main tables use $r=64$), scaling factor $\alpha=2$, dropout 0.0, and no bias adaptation. Adapters are inserted into all attention and MLP projection modules: \texttt{q\_proj}, \texttt{k\_proj}, \texttt{v\_proj}, \texttt{o\_proj}, \texttt{gate\_proj}, \texttt{up\_proj}, and \texttt{down\_proj}.

\paragraph{Training hyperparameters.} We train for 3 epochs with learning rate $1\times 10^{-4}$, with optimizer AdamW, warmup ratio 0.01, and no weight decay. Per-device batch sizes are 16 (train) and 8 (eval) with gradient accumulation of 2, giving an effective train batch size of 32. Sequence length is capped at 256 tokens.

\paragraph{Gradient estimation and projection.} For \method, we compute current and previous gradients with max 8 accumulation steps (script default). Previous gradients include all previous tasks; we build a dataloader per retain task, each using the training batch size. Projection uses a single global coefficient shared across modules.

\paragraph{SVD.} For SVD-based we use randomized low-rank SVD, with $q=4r$ with 4 power iterations.

\paragraph{Evaluation.} We evaluate seen tasks with max. input length 512, using up to 64 evaluation samples per task.

\paragraph{Hardware.} Experiments were run on two NVIDIA RTX A6000 GPUs with 49Gb of VRAM each.


\section{\textsc{NI-Seq-Opposite}}
\label{app:opp-sequences}
Existing Super-NI sequences used in prior work~\citep{jiang2025unlocking,wang2023olora} do not 
control for inter-task gradient interference, leaving sequences with maximal inter-task gradient interference underrepresented in standard evaluations. 
\textsc{NI-Seq-Opposite} fills this gap: we mine 5-task sequences by exhaustive 
combinatorial search over $\binom{46}{5}$ candidate subsets to minimize mean 
pairwise gradient cosine similarity, defined as:

\begin{equation}
\label{eq:seq-conflict}
    \bar\phi(\mathcal{S})
    = \binom{N}{2}^{-1}
      \sum_{1 \le i < j \le N}
      \frac{\langle g_i,\, g_j \rangle}{\|g_i\|\,\|g_j\|},
\end{equation}
where $\mathcal{S} = (T_1, \dots, T_N)$ is a task sequence and $g_i = \mathrm{vec}(G^{(i)}) \in \mathbb{R}^D$ 
is the gradient of the base model on task $T_i$, averaged over a small calibration set. 
Sequences with $\bar\phi \ll 0$ contain many conflicting task pairs.

\paragraph{Mining procedure.}
We compute $g_i$ for each task $T_i$ in a candidate pool of $M$ tasks by 
accumulating gradients on the frozen base model for a small number of steps.
We then score all $\binom{M}{2}$ pairs once and evaluate every candidate 
$N$-task subset by summing precomputed pairwise scores:
\begin{equation}
    \mathcal{S}^* = \arg\min_{\mathcal{S} \subseteq \mathcal{P},\,
    |\mathcal{S}|=N} \bar\phi(\mathcal{S}).
\end{equation}
We searched a pool of $M = 46$ tasks with $N = 5$, evaluating all 
$\binom{46}{5} = 1{,}370{,}754$ candidate subsets from $\binom{46}{2} = 1{,}035$ 
precomputed pair scores.

\paragraph{Selected sequences.}
Table~\ref{tab:ni_seq_opposite_tasks} lists the five adversarial sequences produced by this
search.
The experiments reported in the main manuscript use \textsc{NI-Seq-Opposite-1}, \textsc{NI-Seq-Opposite-2}, and \textsc{NI-Seq-Opposite-3}.

\begin{table}[!t]
\centering
\caption{%
  Task composition of the five \textsc{NI-Seq-Opposite} sequences.
  Sequences 1--3 are used in the main manuscript experiments; 4 and 5 are reported here for completeness.
}
\label{tab:ni_seq_opposite_tasks}
\small
\setlength{\tabcolsep}{3pt}
\resizebox{\linewidth}{!}{%
\begin{tabular}{cll}
\toprule
Seq.\ & Task order (T1 $\to$ T5) & Task IDs \\
\midrule
Opp1 & Typo Verif.\ $\to$ Algebra $\to$ Eval.\ Rel.\ $\to$ SST-2 $\to$ Mutual
  & NI088, NI090, NI1510, NI363, NI611 \\
Opp2 & Typo Verif.\ $\to$ Eval.\ Rel.\ $\to$ Outcome NER $\to$ Set Intersect.\ $\to$ SST-2
  & NI088, NI1510, NI181, NI243, NI363 \\
Opp3 & Typo Verif.\ $\to$ Eval.\ Rel.\ $\to$ Set Intersect.\ $\to$ SST-2 $\to$ Mutual
  & NI088, NI1510, NI243, NI363, NI611 \\
Opp4 & Typo Verif.\ $\to$ Odd-Man-Out $\to$ Outcome NER $\to$ Set Intersect.\ $\to$ SST-2
  & NI088, NI141, NI181, NI243, NI363 \\
Opp5 & Typo Verif.\ $\to$ Eval.\ Sem.\ $\to$ Outcome NER $\to$ Set Intersect.\ $\to$ SST-2
  & NI088, NI1429, NI181, NI243, NI363 \\
\bottomrule
\end{tabular}}
\end{table}

\paragraph{Standard Super-NI sequences.}
For completeness, Table~\ref{tab:ni_seq_standard_tasks} lists the task compositions of the
two standard Super-NI sequences used in the main manuscript, G1 and G2.  Both are pure-generation
sequences drawn from the same Super-NI pool as the adversarial sequences; they are constructed
by grouping tasks by output type without any constraint on gradient alignment.

\begin{table}[!b]
\centering
\caption{%
  Task composition of the two standard Super-NI sequences.
  Both are pure-generation sequences following the construction of Jiang et al.~\citep{jiang2025unlocking}.
}
\label{tab:ni_seq_standard_tasks}
\small
\setlength{\tabcolsep}{3pt}
\resizebox{\linewidth}{!}{%
\begin{tabular}{lll}
\toprule
Seq.\ & Task order (T1 $\to$ T5) & Task IDs \\
\midrule
G1 & Amazon Review $\to$ XSum $\to$ Amazon Food $\to$ Reddit TIFU $\to$ XLSum
   & NI618, NI1290, NI589, NI511, NI1357 \\
G2 & Sent.\ Compression $\to$ Odd-Man-Out $\to$ OhSUMED $\to$ Synth.\ Exec $\to$ Quoref
   & NI1355, NI141, NI619, NI163, NI002 \\
\bottomrule
\end{tabular}}
\end{table}

\paragraph{TRACE sequence.}
Table~\ref{tab:trace_tasks} lists the six tasks comprising the TRACE benchmark
sequence~\citep{wang2023tracecomprehensivebenchmarkcontinual}, which spans diverse domains and task types explicitly
chosen to exhibit interference between sequential fine-tuning stages.

\begin{table}[t]
\centering
\caption{%
  Task composition of the TRACE benchmark sequence.
  Tasks span Chinese and English, covering stance detection, finance, summarization, code, science QA, and math reasoning.
}
\label{tab:trace_tasks}
\small
\setlength{\tabcolsep}{4pt}
\begin{tabular}{llll}
\toprule
Stage & Task & Category & Language \\
\midrule
T1 & C-STANCE    & Stance Detection (MC)  & zh \\
T2 & FOMC        & Finance (MC)           & en \\
T3 & MeetingBank & Summarization          & en \\
T4 & Py150       & Code Generation        & Python \\
T5 & ScienceQA   & Science QA (MC)        & en \\
T6 & NumGLUE-cm  & Math Reasoning         & en \\
\bottomrule
\end{tabular}
\end{table}

\section{Metrics}
\label{appendix:cl-metrics}

We consider a sequence of $T$ tasks $\{t_1, \dots, t_T\}$ learned in order.
Let $R_{i,j}$ denote the score on task $t_j$ after the model has been
sequentially trained on tasks $t_1, \dots, t_i$. These scores form a
lower-triangular matrix where entry $R_{i,j}$ is defined for $j \leq i$.

\paragraph{Average Performance (\map).}
The mean diagonal score, capturing how well the model performs on each task
immediately after learning it:
\begin{equation}
    \mathrm{AP} = \frac{1}{T} \sum_{i=1}^{T} R_{i,i}\,.
\end{equation}

\paragraph{Final Performance (\mfp).}
The mean score over all learned tasks evaluated at the final stage:
\begin{equation}
    \mathrm{FP} = \frac{1}{T} \sum_{j=1}^{T} R_{T,j}\,.
\end{equation}

\paragraph{Forgetting (\mfgt).}
The average drop in per-task performance between the time a task was first
learned and the end of training:
\begin{equation}
    \mathrm{Forget} = \mathrm{AP} - \mathrm{FP}
                    = \frac{1}{T} \sum_{i=1}^{T} \bigl(R_{i,i} - R_{T,i}\bigr)\,.
\end{equation}
A positive value indicates catastrophic forgetting; a negative value indicates that subsequent training improved performance on earlier tasks
(backward transfer).

\paragraph{General Performance (\mgp).}
Let $\mathcal{B} = \{b_1, \dots, b_K\}$ be a set of general-purpose benchmarks
evaluated via \texttt{lm-eval-harness}~\citep{eval-harness}.
Each benchmark $b_k$ is evaluated in a \emph{zero-shot} setting and scored by
its primary metric (accuracy or normalized accuracy when available,
falling back to exact match, F1, or ROUGE-L).
\mgp is the mean score across all benchmarks after the final training stage:
\begin{equation}
    \mathrm{GP} = \frac{1}{K} \sum_{k=1}^{K} s_{b_k}^{(0)}\,,
\end{equation}
where $s_{b_k}^{(0)}$ denotes the zero-shot score on benchmark $b_k$.

\paragraph{In Context Performance (\mip).}
\mip mirrors \mgp but evaluates each benchmark in a \emph{few-shot} setting to
measure the model's ability to follow in-context demonstrations after continual
learning. Concretely, each benchmark is re-evaluated with $n$ in-context
examples ($n{=}5$ for most tasks; $n{=}3$ for BBH tasks):
\begin{equation}
    \mathrm{IP} = \frac{1}{K} \sum_{k=1}^{K} s_{b_k}^{(n_k)}\,,
\end{equation}
where $s_{b_k}^{(n_k)}$ is the $n_k$-shot score on benchmark $b_k$.
The gap $\mathrm{IP} - \mathrm{GP}$ reflects how effectively the model
leverages in-context examples; a shrinking gap signals that the model derives progressively less benefit from in-context demonstrations.

\section{Rank-128 Experiments}
\label{app:rank128exp}

All main manuscript results use LoRA rank $r{=}64$. To assess whether the gains from \method persist at higher rank, we re-run the full comparison at $r{=}128$ across all six evaluation sequences.  
The experimental setup is otherwise identical to the rank-64 protocol.

Table~\ref{tab:r128_cl_metrics} reports \map, \mfp, and \mfgt in the same baseline-plus-delta format as Table~\ref{tab:iclr_delta_all_r64} in the main text.
Table~\ref{tab:r128_gp_ip} reports GP and IP preservation.

\paragraph{Summary of findings.}
The pattern of results at rank 128 closely mirrors rank 64. On adversarial sequences (Opp1 and Opp3), \method consistently recovers large FP gaps relative to Vanilla LoRA: on Opp1, \method ($c{=}1.0$) improves FP by $+20.31$ points and reduces Fgt by $19.34$ points, while on Opp3, \method ($c{=}0.75$) improves FP by $+24.10$ points and reduces Fgt by $19.68$ points. On G2, all three \method variants improve FP by $12$--$15$ points over Vanilla LoRA, consistent with rank-64 results. On standard sequences (G1, TRACE), differences remain small. Finally, \mgp and \mip degradation at rank 128 is comparable to rank 64, remaining close to the Vanilla LoRA reference across most sequences (Table~\ref{tab:r128_gp_ip}).

\begin{table*}[!t]
\centering
\caption{%
  CL metrics across all sequences at rank 128.
  Baseline absolute scores appear in bold rows; indented rows show $\Delta$ for each
  \method variant relative to its base initializer.
}
\label{tab:r128_cl_metrics}
\scriptsize
\setlength{\tabcolsep}{2pt}
\resizebox{\linewidth}{!}{%
\begin{tabular}{l|ccc|ccc|ccc|ccc|ccc|ccc}
\toprule
 & \multicolumn{3}{c}{G1} & \multicolumn{3}{c}{G2} & \multicolumn{3}{c}{TRACE}
 & \multicolumn{3}{c}{Opp1} & \multicolumn{3}{c}{Opp2} & \multicolumn{3}{c}{Opp3} \\
\cmidrule(l){2-19}
Method
  & AP $\uparrow$ & FP $\uparrow$ & Fgt $\downarrow$
  & AP $\uparrow$ & FP $\uparrow$ & Fgt $\downarrow$
  & AP $\uparrow$ & FP $\uparrow$ & Fgt $\downarrow$
  & AP $\uparrow$ & FP $\uparrow$ & Fgt $\downarrow$
  & AP $\uparrow$ & FP $\uparrow$ & Fgt $\downarrow$
  & AP $\uparrow$ & FP $\uparrow$ & Fgt $\downarrow$ \\
\midrule
Vanilla LoRA
  & 9.78 & 9.05 & 0.73
  & 3.12 & 18.64 & $-$15.52
  & 5.35 & 9.50 & $-$4.15
  & 26.39 & 5.33 & 21.07
  & 15.62 & 26.19 & $-$10.57
  & 7.64 & 3.40 & 4.24 \\
\textit{\method ($c{=}0.50$)}
  & \textcolor{ForestGreen}{\textbf{+0.22}} & \textcolor{ForestGreen}{\textbf{+0.21}} & \textcolor{BrickRed}{+0.00}
  & \textcolor{ForestGreen}{\textbf{+0.48}} & \textcolor{ForestGreen}{\textbf{+15.42}} & \textcolor{ForestGreen}{\textbf{$-$14.94}}
  & \textcolor{ForestGreen}{\textbf{+0.47}} & \textcolor{ForestGreen}{\textbf{+4.82}} & \textcolor{ForestGreen}{\textbf{$-$4.35}}
  & \textcolor{ForestGreen}{\textbf{+0.71}} & \textcolor{ForestGreen}{\textbf{+16.99}} & \textcolor{ForestGreen}{\textbf{$-$16.28}}
  & \textcolor{ForestGreen}{\textbf{+10.23}} & \textcolor{ForestGreen}{\textbf{+0.34}} & \textcolor{BrickRed}{+9.89}
  & \textcolor{ForestGreen}{\textbf{+4.36}} & \textcolor{ForestGreen}{\textbf{+23.69}} & \textcolor{ForestGreen}{\textbf{$-$19.33}} \\
\textit{\method ($c{=}0.75$)}
  & \textcolor{ForestGreen}{\textbf{+0.12}} & \textcolor{BrickRed}{+0.00} & \textcolor{BrickRed}{+0.11}
  & \textcolor{ForestGreen}{\textbf{+0.67}} & \textcolor{ForestGreen}{\textbf{+13.87}} & \textcolor{ForestGreen}{\textbf{$-$13.20}}
  & \textcolor{ForestGreen}{\textbf{+7.71}} & \textcolor{ForestGreen}{\textbf{+5.53}} & \textcolor{BrickRed}{+2.17}
  & \textcolor{BrickRed}{$-$0.12} & \textcolor{ForestGreen}{\textbf{+17.22}} & \textcolor{ForestGreen}{\textbf{$-$17.34}}
  & \textcolor{ForestGreen}{\textbf{+9.48}} & \textcolor{ForestGreen}{\textbf{+4.02}} & \textcolor{BrickRed}{+5.46}
  & \textcolor{ForestGreen}{\textbf{+4.42}} & \textcolor{ForestGreen}{\textbf{+24.10}} & \textcolor{ForestGreen}{\textbf{$-$19.68}} \\
\textit{\method ($c{=}1.00$)}
  & \textcolor{ForestGreen}{\textbf{+0.46}} & \textcolor{BrickRed}{$-$0.15} & \textcolor{BrickRed}{+0.60}
  & \textcolor{ForestGreen}{\textbf{+0.78}} & \textcolor{ForestGreen}{\textbf{+12.61}} & \textcolor{ForestGreen}{\textbf{$-$11.83}}
  & \textcolor{BrickRed}{$-$1.29} & \textcolor{ForestGreen}{\textbf{+4.47}} & \textcolor{ForestGreen}{\textbf{$-$5.76}}
  & \textcolor{ForestGreen}{\textbf{+0.96}} & \textcolor{ForestGreen}{\textbf{+20.31}} & \textcolor{ForestGreen}{\textbf{$-$19.34}}
  & \textcolor{ForestGreen}{\textbf{+9.55}} & \textcolor{ForestGreen}{\textbf{+4.39}} & \textcolor{BrickRed}{+5.17}
  & \textcolor{BrickRed}{$-$2.71} & \textcolor{ForestGreen}{\textbf{+23.22}} & \textcolor{ForestGreen}{\textbf{$-$25.93}} \\
\midrule
LoRAM
  & 9.68 & 8.88 & 0.79
  & 4.73 & 15.45 & $-$10.72
  & 3.36 & 11.46 & $-$8.10
  & 25.96 & 24.82 & 1.15
  & 27.73 & 30.05 & $-$2.31
  & 8.62 & 28.08 & $-$19.46 \\
\textit{\method ($c{=}0.50$)}
  & \textcolor{ForestGreen}{\textbf{+0.33}} & \textcolor{ForestGreen}{\textbf{+0.38}} & \textcolor{ForestGreen}{\textbf{$-$0.05}}
  & \textcolor{BrickRed}{$-$1.13} & \textcolor{ForestGreen}{\textbf{+18.61}} & \textcolor{ForestGreen}{\textbf{$-$19.73}}
  & \textcolor{ForestGreen}{\textbf{+2.46}} & \textcolor{ForestGreen}{\textbf{+2.87}} & \textcolor{ForestGreen}{\textbf{$-$0.41}}
  & \textcolor{ForestGreen}{\textbf{+1.14}} & \textcolor{BrickRed}{$-$2.50} & \textcolor{BrickRed}{+3.65}
  & \textcolor{BrickRed}{$-$1.88} & \textcolor{BrickRed}{$-$3.51} & \textcolor{BrickRed}{+1.64}
  & \textcolor{ForestGreen}{\textbf{+3.38}} & \textcolor{BrickRed}{$-$0.98} & \textcolor{BrickRed}{+4.36} \\
\textit{\method ($c{=}0.75$)}
  & \textcolor{ForestGreen}{\textbf{+0.23}} & \textcolor{ForestGreen}{\textbf{+0.17}} & \textcolor{BrickRed}{+0.05}
  & \textcolor{BrickRed}{$-$0.94} & \textcolor{ForestGreen}{\textbf{+17.06}} & \textcolor{ForestGreen}{\textbf{$-$18.00}}
  & \textcolor{ForestGreen}{\textbf{+9.69}} & \textcolor{ForestGreen}{\textbf{+3.58}} & \textcolor{BrickRed}{+6.12}
  & \textcolor{ForestGreen}{\textbf{+0.31}} & \textcolor{BrickRed}{$-$2.27} & \textcolor{BrickRed}{+2.58}
  & \textcolor{BrickRed}{$-$2.63} & \textcolor{ForestGreen}{\textbf{+0.16}} & \textcolor{ForestGreen}{\textbf{$-$2.79}}
  & \textcolor{ForestGreen}{\textbf{+3.44}} & \textcolor{BrickRed}{$-$0.57} & \textcolor{BrickRed}{+4.01} \\
\textit{\method ($c{=}1.00$)}
  & \textcolor{ForestGreen}{\textbf{+0.56}} & \textcolor{ForestGreen}{\textbf{+0.02}} & \textcolor{BrickRed}{+0.54}
  & \textcolor{BrickRed}{$-$0.83} & \textcolor{ForestGreen}{\textbf{+15.79}} & \textcolor{ForestGreen}{\textbf{$-$16.62}}
  & \textcolor{ForestGreen}{\textbf{+0.70}} & \textcolor{ForestGreen}{\textbf{+2.51}} & \textcolor{ForestGreen}{\textbf{$-$1.81}}
  & \textcolor{ForestGreen}{\textbf{+1.39}} & \textcolor{ForestGreen}{\textbf{+0.82}} & \textcolor{BrickRed}{+0.58}
  & \textcolor{BrickRed}{$-$2.55} & \textcolor{ForestGreen}{\textbf{+0.53}} & \textcolor{ForestGreen}{\textbf{$-$3.09}}
  & \textcolor{BrickRed}{$-$3.70} & \textcolor{BrickRed}{$-$1.46} & \textcolor{ForestGreen}{\textbf{$-$2.23}} \\
\midrule
LoRA-GA
  & 10.08 & 9.09 & 0.99
  & 4.56 & 31.89 & $-$27.33
  & 15.40 & 14.87 & 0.53
  & 26.56 & 26.62 & $-$0.06
  & 24.81 & 26.24 & $-$1.43
  & 10.27 & 18.00 & $-$7.73 \\
\textit{\method ($c{=}0.50$)}
  & \textcolor{BrickRed}{$-$0.08} & \textcolor{ForestGreen}{\textbf{+0.18}} & \textcolor{ForestGreen}{\textbf{$-$0.26}}
  & \textcolor{BrickRed}{$-$0.96} & \textcolor{ForestGreen}{\textbf{+2.16}} & \textcolor{ForestGreen}{\textbf{$-$3.12}}
  & \textcolor{BrickRed}{$-$9.58} & \textcolor{BrickRed}{$-$0.55} & \textcolor{ForestGreen}{\textbf{$-$9.03}}
  & \textcolor{ForestGreen}{\textbf{+0.55}} & \textcolor{BrickRed}{$-$4.31} & \textcolor{BrickRed}{+4.85}
  & \textcolor{ForestGreen}{\textbf{+1.04}} & \textcolor{ForestGreen}{\textbf{+0.29}} & \textcolor{BrickRed}{+0.75}
  & \textcolor{ForestGreen}{\textbf{+1.73}} & \textcolor{ForestGreen}{\textbf{+9.10}} & \textcolor{ForestGreen}{\textbf{$-$7.37}} \\
\textit{\method ($c{=}0.75$)}
  & \textcolor{BrickRed}{$-$0.18} & \textcolor{BrickRed}{$-$0.03} & \textcolor{ForestGreen}{\textbf{$-$0.15}}
  & \textcolor{BrickRed}{$-$0.78} & \textcolor{ForestGreen}{\textbf{+0.61}} & \textcolor{ForestGreen}{\textbf{$-$1.39}}
  & \textcolor{BrickRed}{$-$2.34} & \textcolor{ForestGreen}{\textbf{+0.16}} & \textcolor{ForestGreen}{\textbf{$-$2.51}}
  & \textcolor{BrickRed}{$-$0.29} & \textcolor{BrickRed}{$-$4.08} & \textcolor{BrickRed}{+3.79}
  & \textcolor{ForestGreen}{\textbf{+0.30}} & \textcolor{ForestGreen}{\textbf{+3.97}} & \textcolor{ForestGreen}{\textbf{$-$3.67}}
  & \textcolor{ForestGreen}{\textbf{+1.79}} & \textcolor{ForestGreen}{\textbf{+9.51}} & \textcolor{ForestGreen}{\textbf{$-$7.72}} \\
\textit{\method ($c{=}1.00$)}
  & \textcolor{ForestGreen}{\textbf{+0.16}} & \textcolor{BrickRed}{$-$0.18} & \textcolor{BrickRed}{+0.34}
  & \textcolor{BrickRed}{$-$0.66} & \textcolor{BrickRed}{$-$0.65} & \textcolor{ForestGreen}{\textbf{$-$0.01}}
  & \textcolor{BrickRed}{$-$11.34} & \textcolor{BrickRed}{$-$0.90} & \textcolor{ForestGreen}{\textbf{$-$10.44}}
  & \textcolor{ForestGreen}{\textbf{+0.80}} & \textcolor{BrickRed}{$-$0.98} & \textcolor{BrickRed}{+1.78}
  & \textcolor{ForestGreen}{\textbf{+0.37}} & \textcolor{ForestGreen}{\textbf{+4.34}} & \textcolor{ForestGreen}{\textbf{$-$3.97}}
  & \textcolor{BrickRed}{$-$5.34} & \textcolor{ForestGreen}{\textbf{+8.62}} & \textcolor{ForestGreen}{\textbf{$-$13.96}} \\
\bottomrule
\end{tabular}
}
\end{table*}

\begin{table}[!b]
\centering
\caption{%
  GP and IP preservation at rank 128.  The first row reports absolute scores for
  Vanilla LoRA; all remaining rows show $\Delta$ relative to Vanilla LoRA.
  \method variants remain close to the reference on GP
  across all sequences.
}
\label{tab:r128_gp_ip}
\scriptsize
\setlength{\tabcolsep}{3pt}
\resizebox{\linewidth}{!}{%
\begin{tabular}{l|cc|cc|cc|cc|cc|cc}
\toprule
 & \multicolumn{2}{c}{G1} & \multicolumn{2}{c}{G2} & \multicolumn{2}{c}{TRACE}
 & \multicolumn{2}{c}{Opp1} & \multicolumn{2}{c}{Opp2} & \multicolumn{2}{c}{Opp3} \\
\cmidrule(l){2-13}
Method
  & $\Delta$GP & $\Delta$IP
  & $\Delta$GP & $\Delta$IP
  & $\Delta$GP & $\Delta$IP
  & $\Delta$GP & $\Delta$IP
  & $\Delta$GP & $\Delta$IP
  & $\Delta$GP & $\Delta$IP \\
\midrule
Vanilla LoRA (ref.)
  & 50.79 & 59.14
  & 51.97 & 58.70
  & 55.40 & 56.34
  & 51.70 & 59.35
  & 55.56 & 61.49
  & 51.89 & 59.71 \\
\midrule
LoRAM
  & $-$0.31 & $-$2.38
  & $-$0.85 & $-$5.12
  & $-$2.37 & $+$0.02
  & $-$0.49 & $-$5.24
  & $-$1.36 & $-$1.76
  & $-$0.88 & $-$2.41 \\
LoRA-GA
  & $-$0.10 & $-$2.28
  & $-$1.34 & $-$2.58
  & $-$1.77 & $-$1.87
  & $-$1.21 & $-$4.28
  & $-$2.22 & $-$2.95
  & $-$0.60 & $-$4.50 \\
\midrule
SLICE ($c{=}0.50$)
  & $-$0.09 & $-$2.29
  & $-$1.43 & $-$3.48
  & $-$1.44 & $-$6.92
  & $-$1.26 & $-$4.84
  & $-$2.28 & $-$2.10
  & $-$0.27 & $-$2.86 \\
SLICE ($c{=}0.75$)
  & $-$0.18 & $-$2.24
  & $-$1.45 & $-$1.90
  & $-$2.03 & $-$3.02
  & $-$0.66 & $-$3.67
  & $-$2.45 & $-$1.40
  & $-$0.16 & $-$2.36 \\
SLICE ($c{=}1.00$)
  & $+$0.23 & $-$2.33
  & $-$1.59 & $-$4.34
  & $-$1.72 & $-$2.04
  & $-$0.79 & $-$4.97
  & $-$2.12 & $-$2.24
  & $-$0.58 & $-$2.71 \\
\bottomrule
\end{tabular}
}
\end{table}

\section{Additional Performance Heatmaps on \textsc{NI-Seq-Opposite-1}}
\label{app:heatmaps}

\begin{figure}[t]
    \centering
    \includegraphics[width=\linewidth]{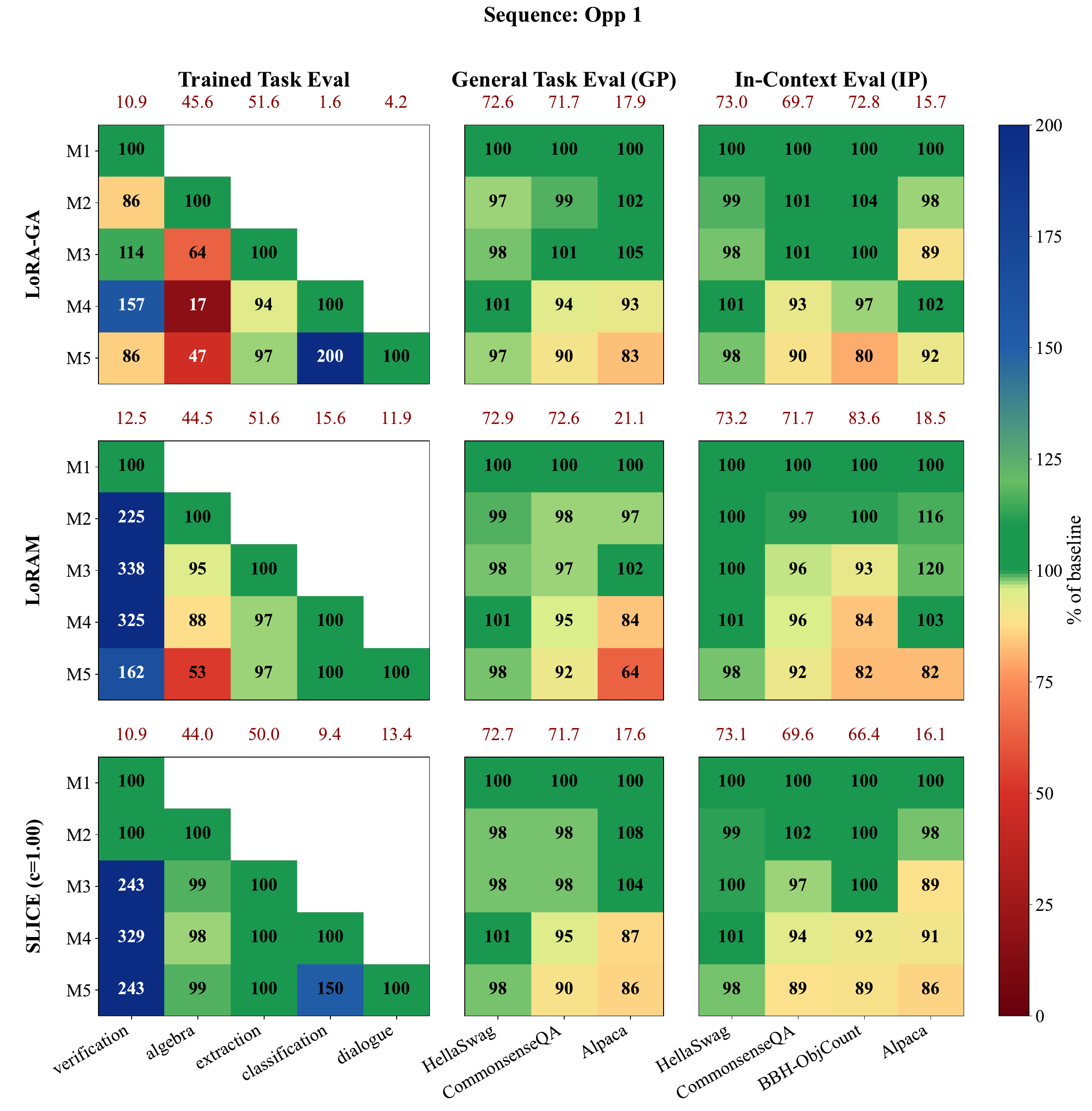}
    \caption{\textbf{Performance heatmaps comparing LoRA-GA vs.\ LoRAM vs.\ \method ($c=1.0$)\ during continual learning on \textsc{NI-Seq-Opposite-1}.}
    Numbers above each heatmap indicate baseline performance on the trained task, \mgp and \mip, at left, center, and right, respectively. 
    Heatmap values show percentage change relative to baseline.}
    \label{fig:opp1_lora_ga_vs_slice}
\end{figure}

Figure~\ref{fig:opp1-comparison} in the main manuscript compares Vanilla LoRA against \method ($c{=}1.0$) on \textsc{NI-Seq-Opposite-1}.
Vanilla LoRA is chosen as the reference there because it achieves the highest \map among the baselines on this sequence, making the forgetting contrast the most visually striking.

Figure~\ref{fig:opp1_lora_ga_vs_slice} shows the analogous three-panel heatmaps for the two remaining baselines, LoRA-GA and LoRAM, paired against \method ($c{=}1.0$) in each case.
The layout is identical to Figure~\ref{fig:opp1-comparison}: the left panel shows per-task trained-task evaluation (values are percentage of each task's baseline score, so 100 = parity with the model before any continual training), the center panel shows zero-shot \mgp on the four held-out benchmarks, and the right panel shows few-shot \mip on the same benchmarks plus BBH Object Counting.

\paragraph{LoRA-GA vs.\ \method.}
LoRA-GA initializes adapters using only the current-task gradient, achieving higher \map than Vanilla LoRA on \textsc{NI-Seq-Opposite-1} due to its task-aligned initialization.
However, its off-diagonal entries still decay substantially across stages: tasks learned early in the sequence suffer significant forgetting by the final stage.
\method ($c{=}1.0$) recovers or improves upon LoRA-GA's off-diagonal retention while matching its diagonal scores, confirming that the gradient-surgery projection adds value beyond the task-aware subspace that LoRA-GA already captures.
\mgp and \mip degradation are comparable between the two methods.

\paragraph{LoRAM vs.\ \method.}
LoRAM initialization relies on a deterministic orthogonal basis constructed via the Discrete Sine Transform.
LoRAM achieves a strong AP on this sequence. 
However, like LoRA-GA, its task-agnostic subspace selection leaves it vulnerable to inter-task interference: the below-diagonal entries decay noticeably across stages.
\method ($c{=}1.0$) achieves better off-diagonal retention throughout the sequence, reducing forgetting while keeping AP competitive. 
The modest \mgp and \mip reductions are consistent with the aggregate results in Table~\ref{tab:gp_ip_all_r64} of the main manuscript.



\end{document}